\documentclass[preprint,12pt]{elsarticle}

\usepackage{amssymb}
\usepackage{amsmath}

\usepackage{hyperref}   
\usepackage[utf8]{inputenc} 
\usepackage[T1]{fontenc}    

\usepackage{url}            
\usepackage{booktabs}       
\usepackage{amsfonts}       
\usepackage{nicefrac}       
\usepackage{microtype}      
\usepackage{xcolor}         

\usepackage{cleveref}
\usepackage{subfigure}
\usepackage{graphicx}
\usepackage{listings}
\usepackage{subcaption}
\usepackage{fancyvrb}
\usepackage{float}
\usepackage{multirow}

\newcommand{\blue}[1]{\textcolor{black}{#1}}
\PassOptionsToPackage{capitalize,noabbrev}{cleveref}
\usepackage{cleveref}
\crefrangelabelformat{table}{#3#1-#2#4} 

\journal{Neurocomputing}

\begin{document}

\begin{frontmatter}



\title{PointT2I: LLM-based text-to-image generation via keypoints}


\author[IPAI]{Taekyung Lee\texorpdfstring{\fnref{label1}}{}}
\ead{dlxorud1231@snu.ac.kr}
\author[math]{Donggyu Lee\texorpdfstring{\fnref{label1}}{}}
\ead{dglee442@snu.ac.kr}
\author[math]{Myungjoo Kang\texorpdfstring{\corref{cor1}}{}}
\ead{mkang@snu.ac.kr}
\cortext[cor1]{corresponding author}
\fntext[label1]{equal contribution}



\affiliation[IPAI]{organization={Interdisciplinary Program in Artificial Intelligence, Seoul National University},
          addressline={1 Gwanak-ro, Gwanak-gu}, 
            city={Seoul},
            postcode={08826}, 
            country={South Korea}}
\affiliation[math]{organization={Department of Mathematical Sciences, Seoul National University},
            addressline={1 Gwanak-ro, Gwanak-gu}, 
            city={Seoul},
            postcode={08826}, 
            country={South Korea}}

\begin{abstract}
Text-to-image (T2I) generation model has made significant advancements, resulting in high-quality images aligned with an input prompt. However, despite T2I generation's ability to generate fine-grained images, it still faces challenges in accurately generating images when the input prompt contains complex concepts, especially human pose.
In this paper, we propose \textbf{PointT2I}, a framework that effectively generates images that accurately correspond to the human pose described in the prompt by using a large language model (LLM). 
 \textbf{PointT2I} consists of three components: Keypoint generation, Image generation, and Feedback system. The keypoint generation uses an LLM to directly generate keypoints corresponding to a human pose, solely based on the input prompt, without external references. Subsequently, the image generation produces images based on both the text prompt and the generated keypoints to accurately reflect the target pose. To refine the outputs of the preceding stages, we incorporate an LLM-based feedback system that assesses the semantic consistency between the generated contents and the given prompts. Our framework is the first approach to leveraging LLM for keypoints-guided image generation without any fine-tuning, producing accurate pose-aligned images based solely on textual prompts. 
\end{abstract}






\end{frontmatter}



\section{Introduction}

With the emergence of diffusion models, text-to-image (T2I) generation has seen substantial progress, exemplified by Stable Diffusion (SD) and Stable Diffusion XL (SDXL) \citep{sd, sdxl}. Most T2I models \citep{sd, sdxl, saharia2022photorealistic, ruiz2023dreambooth} rely solely on natural language prompts and lack the capacity to incorporate other input modalities. This reliance on prompts alone imposes clear limitations, making it difficult to develop models capable of controllable and accurate human pose generation \citep{controlnet, t2i}. An effective strategy to address this limitation is to incorporate auxiliary inputs that are semantically aligned with the input prompts. Several methods---such as ControlNet \cite{controlnet}, T2I-Adapter \cite{t2i}, GLIGEN \cite{gligen}, and HumanSD \cite{humansd}---extend conventional T2I generation by accepting additional modalities such as keypoints or pose skeleton images. Although these approaches enable more precise control over attributes---such as human body pose and object placement---during image generation, they inherently rely on auxiliary inputs derived from external data sources. Such reliance leads to challenges when these external sources are unavailable or misaligned with the intended meaning of the input prompt.

To address this limitation, it is desirable to develop methods that generate additional inputs directly from the prompt, rather than from external data. Recent studies \citep{feng2023layoutgpt, lian2023llm, wu2024self, gani2023llm} have shown that Large Language Models (LLMs) can effectively capture the semantics of textual prompts, leading to improved alignment between language input and visual outputs in T2I models. For instance, LayoutGPT \cite{feng2023layoutgpt} leverages an LLM to generate scene layouts from prompts, explicitly defining object positions and spatial arrangements. Building on this, several studies \citep{lian2023llm, wu2024self, gani2023llm} incorporate LLM-generated scene layouts into the image generation process. The use of LLM-generated scene layouts allows the generation model to more accurately capture structural attributes of the prompt, such as the number, position, and spatial arrangement of objects. While most existing studies have focused on relatively simple attributes such as object count and position, we address the more challenging task of human pose generation. By leveraging LLM, we aim to interpret pose-specific descriptions in prompts and generate images that accurately reflect fine-grained body configurations.

To this end, we propose \textbf{PointT2I}, a framework that infers human poses solely from input textual prompts, without relying on external auxiliary data or any fine-tuning for pose representation. \blue{Rather than modifying or extending the internal control mechanisms of pose-aware T2I backbones, such as ControlNet and HumanSD, PointT2I serves as a semantic-to-structural translator that unlocks their latent potential. While these backbones can generate human-pose images conditioned on structural inputs---such as pose keypoints or skeleton maps---they cannot directly interpret such structures from textual prompts.}

\blue{To address this limitation,} PointT2I first predicts human pose keypoints from the input text prompt using an LLM. These keypoints provide structural guidance that unlocks the pose-handling capacity of pose-aware T2I backbones. This makes it possible to generate images that accurately reflect the pose described in the prompt. To improve alignment between the generated image and the pose described in the prompt, PointT2I employs a two-module feedback system based on LLM reasoning. The first module refines the predicted keypoints, and the second adjusts the generated image to better match the intended pose. This design allows PointT2I to act as a semantic-to-structural bridge: it translates high-level language descriptions into concrete pose structures that pose-aware backbones can render.
As a result, PointT2I can handle diverse human poses, including uncommon and challenging ones such as yoga and acrobatics. Importantly, all of these capabilities are achieved without any pose-specific training or adaptation, using only a frozen LLM. In addition, PointT2I remains robust to different types of pose descriptions. It accurately handles both explicit pose names (e.g., “performing the boat pose”) and descriptive instructions (e.g., “holding a V-shaped pose with raised legs and leaned-back torso”), consistently generating pose-accurate images.

Through extensive experiments, we demonstrate that PointT2I effectively generates pose-accurate images across a diverse range of human actions, including challenging and uncommon poses. The results also highlight the effectiveness of the LLM-based feedback system in refining both keypoints and images, and confirm the model’s robustness to variations in prompt phrasing and descriptive detail. Our main contributions are as follows:

\begin{itemize}
    \item \textbf{PointT2I is the first framework to leverage LLMs to generate human pose keypoints directly from textual prompts.} This allows the model to act as a semantic-to-structural bridge, unlocking the potential of pose-aware T2I backbones. As a result, it can handle a wide range of human poses without any fine-tuning.
    
    \item \textbf{The LLM-based feedback system improves pose-level consistency by refining both the predicted keypoints and the generated images.} This two-module refinement process enhances alignment between the textual description and the visual output.
    
    \item \textbf{PointT2I exhibits strong robustness to variation in prompt styles, effectively interpreting both explicit pose names and descriptive language.} Accurate image generation is maintained even under diverse phrasings and levels of detail.

\end{itemize}

\label{sec:intro}

\section{Background}
\label{sec:Background}

\subsection{Diffusion-based Text-to-Image Generation}

Diffusion models have emerged as a powerful approach for T2I generation, enabling semantically aligned and visually realistic images from textual descriptions. Among early models, GLIDE \cite{nichol2022glide} introduces classifier-free guidance to better align the generated image with the input text. SD, based on the Latent Diffusion Model \cite{sd}, improves computational efficiency by operating in a compressed latent space, supporting high-resolution synthesis with reduced resource demands. SDXL \cite{sdxl} extends this approach with substantial architectural and performance improvements. It employs a deeper and wider U-Net backbone and incorporates dual text encoders (CLIP \cite{radford2021learning} and T5 \cite{t5}) to improve prompt comprehension.

While effective in generating visually consistent results, these models offer limited controllability \cite{controlnet,t2i}. To control diffusion-based generation models, many studies have incorporated additional external inputs to guide image generation \cite{controlnet,t2i,gligen,yang2023reco}. For example, ReCo \cite{yang2023reco} employs explicit position tokens to control the layout condition. In contrast, ControlNet \cite{controlnet} and T2I-Adapter \cite{t2i} guide the generation process using many types of auxiliary inputs such as canny edges or sketches. ControlNet integrates these conditions into each layer of the U-Net via a duplicated branch, enabling fine-grained control throughout the generation process. T2I-Adapter, on the other hand, attaches lightweight external modules, allowing rapid adaptation with minimal retraining while preserving the base model.

A similar strategy is employed in text-guided human image generation, with body pose commonly represented using auxiliary inputs such as keypoints or skeleton images. HumanSD \cite{humansd}, tailored for human pose synthesis, follows the skeleton-based approach, similar to ControlNet \cite{controlnet} and T2I-Adapter \cite{t2i}. In contrast, GLIGEN \cite{gligen} directly utilizes keypoints as conditioning inputs. Previous methods, including GLIGEN and HumanSD, rely on external conditions for pose \blue{specification}, which limits their applicability. We propose a framework that uses an LLM to generate keypoints from the text prompt, enabling more effective pose \blue{representation} across diverse human poses without additional inputs or fine-tuning.

\subsection{Text-to-Image Generation using LLM}
LLMs, such as the GPT series \cite{gpt2,gpt3} and T5 \cite{t5}, have significantly advanced natural language processing by effectively understanding and generating natural language. These models are trained on extensive text datasets, allowing them to learn intricate linguistic structures and contextual dependencies. Building on their success in language understanding, recent advances employ LLM to guide the T2I generation of diffusion models \cite{lian2023llm,wu2024self,gani2023llm, lin2023videodirectorgpt, qu2023layoutllm}. In this context, two main approaches have emerged: (1) optimizing prompts for better alignment, and (2) generating additional conditions from the prompt, each addressing different aspects of guiding text-to-image generation.

Following the first approach, Promptist \cite{hao2023optimizing} and PAE \cite{mo2024dynamic} optimize prompts using reinforcement learning. Promptist \cite{hao2023optimizing} adapts user prompts into model-preferred forms to improve image quality. PAE \cite{mo2024dynamic} further enhances images by adjusting each word's weight and injection time step in the prompt. The second approach primarily introduces layout-based conditions for more explicit guidance. LMD \cite{lian2023llm} generates captioned bounding boxes from the prompt. LayoutLLM-T2I \cite{qu2023layoutllm} not only generates bounding boxes from prompts, but also optimizes in-context learning via a feedback-based sampler and enhances synthesis through layout-guided interaction. LLM Blueprint \cite{gani2023llm} decomposes complex prompts into smaller components to construct spatial layouts. SLD \cite{wu2024self} further refines layout quality via a closed-loop feedback mechanism. However, since layouts encode only numeric and spatial information, these methods lack the semantic detail required for precise pose guidance in human image generation.

Our method remains effective across a wide range of prompt formulations while maintaining semantic alignment with the described actions. 
The proposed method directly generates keypoints for human pose synthesis without optimizing the prompt or relying on layout, which makes the method fundamentally distinct from both approaches.

\subsection{Broader Multimodal model Research}

Recently, multimodal models have attracted considerable attention and become an active area of research. These models are capable of processing and integrating information from multiple modalities, such as text, images, audio, and video \cite{baltruvsaitis2018multimodal}.
Various multimodal models have seen significant progress in recent years, including image-to-text models, video-based multimodal models, and the aforementioned T2I model, along with other types of multimodal models.

In the image-to-text domain, recent efforts have focused on improving image captioning.
LSTNet \cite{ma2023towards} improves image captioning by refining local attention within each layer and integrating spatial cues across layers to capture contextual details.
DTNet \cite{ma2024image} dynamically selects processing paths based on both spatial and channel features of the input image, enabling more accurate and distinctive captions than static Transformer approaches.

Multimodal research in the video domain has also gained momentum.
Video-of-Thought \cite{fei2024video} adapts Chain-of-Thought reasoning to video understanding, breaking down complex tasks into sequential steps based on spatio-temporal scene graph representations.
Finsta \cite{fei2024enhancing} improves video-language understanding by aligning fine-grained spatio-temporal semantics across modalities using scene graphs and contrastive learning in a plug-and-play framework.
Dysen-VDM \cite{fei2024dysen} utilizes LLMs such as ChatGPT to extract structured action sequences and enrich textual descriptions, which guide fine-grained scene graphs for conditioning video diffusion models.

In addition, more general any-to-any multimodal frameworks have emerged.
NExT-GPT \cite{wu2024next} enables unified multimodal understanding and generation across text, image, video, and audio modalities by combining pretrained encoders/decoders with an LLM and lightweight alignment modules.

Multimodal research has increasingly advanced in both scope and capability. In this paper, we focus on T2I model and propose a novel framework tailored to generating imagery guided by human poses.

\begin{figure*}[t] 
    \centering
        \includegraphics[width=1.0\textwidth]{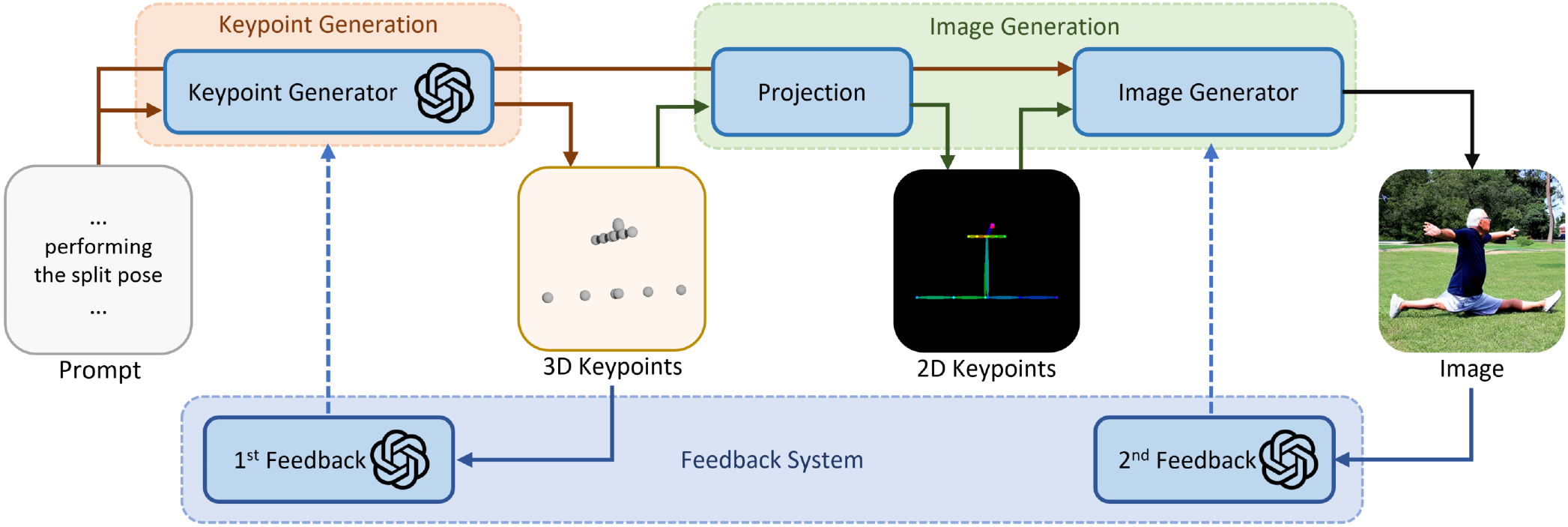}
    \caption{\textbf{The overview of PointT2I}. PointT2I consists of three components: keypoint generation, image generation, and feedback system, designed to produce images that accurately depict the intended pose.
    }
    \label{fig:overall-structure}
    \vspace{-8pt}
\end{figure*}

\section{Method}
\label{sec:method}

Existing LLM-based T2I models \citep{lian2023llm,wu2024self, gani2023llm} enhance prompt-image alignment by generating additional inputs, typically in the form of spatial layouts. However, these approaches are limited to numeric and positional cues, lacking the capacity to capture more expressive semantics such as fine-grained human poses. 

To tackle this limitation, we propose \textbf{PointT2I}, a framework that generates pose-consistent images by guiding a T2I model with human pose keypoints inferred from text via an LLM. Importantly, this process requires no pose-specific fine-tuning. The workflow proceeds as follows: the input prompt is first analyzed to extract pose-related semantics, which are then transformed into keypoints. Based on these keypoints, the model generates an image that aligns with the described pose. A feedback system evaluates both the generated keypoints and the resulting image against the prompt. It detects pose-level inconsistencies---such as mismatches between the described pose and the visual output---and triggers regeneration when necessary. Through this pipeline, our framework effectively produces images that faithfully reflect the prompt and its associated pose. Furthermore, by leveraging the flexibility of LLMs, PointT2I generalizes well to diverse human poses.

\textbf{PointT2I} consists of three sequential components, illustrated in  \Cref{fig:overall-structure}:

\begin{itemize}
    \item[(1)] \textbf{Keypoint Generation:} An LLM interprets the input prompt and predicts a set of three-dimensional (3D) human pose keypoints.
    \item[(2)] \textbf{Image Generation:} A pose-aligned image is generated from the input prompt and two-dimensional (2D) keypoints projected from the predicted 3D pose.   
    \item[(3)] \textbf{Feedback System:} An LLM-based module evaluates both the keypoints and the generated image. If a misalignment with the prompt is detected, it refines the outputs via regeneration.
\end{itemize}

In the following, we detail how each component of PointT2I functions and contributes to generating accurate human pose images.

\subsection{Keypoint Generation} 
PointT2I employs keypoints as guidance for image generation. Therefore, it is crucial to obtain keypoints that are semantically aligned with the input prompt. However, generating prompt-specific keypoints typically requires additional network training, which is costly and impractical in many real-world settings. To overcome this challenge, we leverage an LLM to extract and interpret action-related information from the input prompt and use it to compute 3D keypoint coordinates for 17 major human body joints. Leveraging this LLM-based keypoint inference, our framework performs well on uncommon and challenging poses, as it does not rely on pose-specific training or pose annotations.

To implement this process, the keypoint generator first analyzes the input prompt to identify pose-relevant semantics, such as body configuration and ground-contact cues. It filters out descriptive or contextual content that is not directly related to the pose. Based on the extracted pose description, the generator then determines which body parts are involved and assigns 3D coordinates to 17 major joints. To produce realistic and human-like poses, the generator anchors contact points (e.g., feet, hands) at \(z = 0\) and maintains realistic body proportions across all generated poses.

In summary, the keypoint generator infers 3D human pose keypoints directly from input textual prompts using an LLM, without relying on external auxiliary inputs. It maps action-related segments of the prompt into structured keypoints that define the target pose. This enables the generation of a wide range of poses without any fine-tuning, while preserving human-like body structure. See \ref{code:keypoints} for more details.

\subsection{Image Generation}
To generate pose-guided images, we project the 3D keypoints produced by our keypoint generator into a 2D representation suitable for image synthesis. Among multiple candidate projections, we select the most informative one based on spatial distribution. The selected 2D keypoints, which we refer to as \textit{pose guidance}, are provided to the image generator along with the input prompt to produce pose-consistent images.

We obtain a set of 3D human pose keypoints from the keypoint generator, with $z=0$ denoting the ground plane. \blue{To make them suitable for image generation, we project the 3D keypoints onto a 2D image plane using an orthographic camera model, which linearly maps 3D points onto the image plane by discarding the depth component along the viewing direction.
The camera intrinsics are assumed to have a unit focal length and the image center as the principal point, implying that the original 3D coordinates are preserved without additional scaling or normalization. The camera extrinsics are defined by the view direction, specified as $\mathbf{v}= (\cos (\theta\pi/8), \sin(\theta\pi/8), 0)$ for $\theta = 0,1,..., 7$, corresponding to eight orthographic views uniformly distributed around the vertical axis.}
Among these projections, we select the one with the highest 2D coordinate variance to minimize overlap and better preserve the spatial structure of the original 3D human pose. The selected projection serves as the pose guidance used in the image generation process. Each of the 17 keypoints in this projection corresponds to a specific human joint, facilitating an accurate representation of the target pose. Together with the textual prompt, this projection guides the image generator in producing an image aligned with the specified pose. In \Cref{sec:experiments}, we present both qualitative and quantitative results demonstrating that the generated images align well with the textual input prompts and the generated human pose keypoints.

\begin{figure*}[t] 
    \centering
    \subfigure[First feedback: Feedback module for the generated keypoints.]{
        \includegraphics[width=0.95\textwidth]{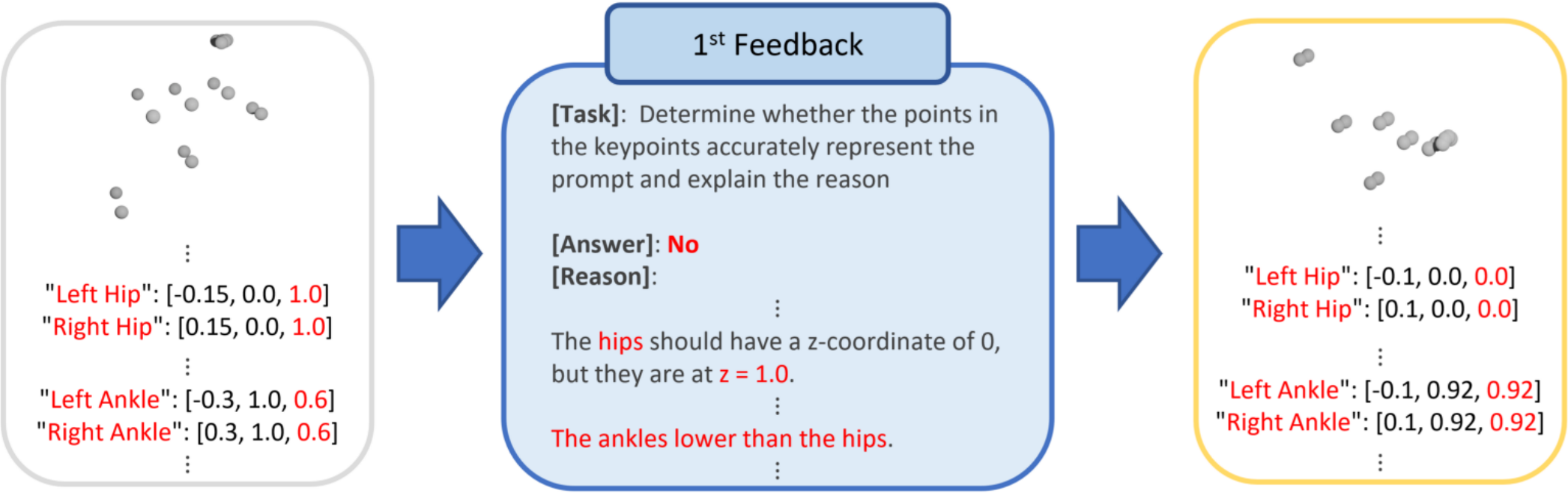}
        \vspace{1mm}
        }
    \subfigure[Second feedback: Feedback module for the generated images.]{
        \includegraphics[width=0.95\textwidth]{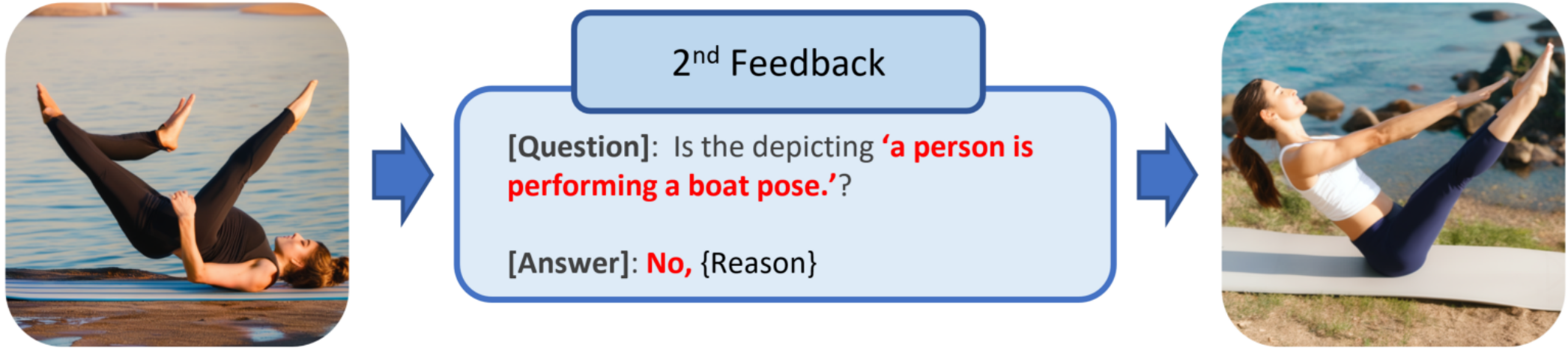}
        }
       
    \caption{\textbf{Feedback System}. The feedback system is composed of two modules, each employing an LLM to assess the intermediate outputs of the generation pipeline. Based on this evaluation, the system provides feedback to the generator, guiding the generator toward improved alignment with the input prompt.
    }
    \label{fig:feedback structure}
    \vspace{-8pt}
\end{figure*}

\subsection{Feedback System}
We propose a two-module feedback system that leverages an LLM to enforce pose-level consistency throughout our framework pipeline. Given a textual prompt, the system refines both the predicted keypoints and the generated image to ensure alignment with the described pose. This process yields images that are better aligned with the prompt without relying on external annotations.

The feedback system operates on both the keypoint and image generators, as illustrated in \Cref{fig:overall-structure}. A more detailed schematic of the feedback pipeline is provided in \Cref{fig:feedback structure}, showing how the two LLM-based modules interact with intermediate outputs. We describe the function of each module below.

The first feedback module operates on the predicted keypoints.  It takes the 3D keypoints and the original prompt as inputs and employs an LLM to assess whether the keypoints accurately reflect the described actions. If inconsistencies such as incorrect joint positions are detected, the first module provides an explanation and generates revised keypoints accordingly. The second module evaluates whether the generated image aligns with the pose described in the input prompt. If misalignment is detected, the module triggers image regeneration to improve consistency with the prompt. This process ensures that the final output is aligned with the prompt and faithfully represents the intended pose.

This LLM-based feedback system enables end-to-end refinement of intermediate outputs without relying on ground-truth pose labels or human supervision. In \Cref{sec:feedback}, we demonstrate that each module of feedback contributes to improved alignment with the input prompt. Examples of the actual LLM templates used in each module are provided in \ref{code:1st_feedback} and \ref{code:2nd_feedback}.

\section{Experiments}
\label{sec:experiments}

In this section, we present empirical evidence demonstrating that our framework achieves competitive performance in depicting human poses. Our approach leverages an LLM to extract keypoints from a prompt describing a specific pose, which are used to guide image generation and ensure alignment with the intended pose. To evaluate the effectiveness of our method, we generate images from prompts covering a wide range of human postures. 
We assess the quality and prompt relevance of the generated images by measuring image–prompt similarity \cite{lin2024evaluating,hessel2021clipscore} and performing classification experiments.

\subsection{Implementation Details}
\label{subsec:imple_detail}
\paragraph{Framework} Our framework consists of three components: keypoint generation, image generation, and a feedback system. For keypoint generation, we use ChatGPT o1-preview \cite{OpenAI2024o1} as a keypoint generator to extract 3D human pose coordinates from textual prompts. These keypoints are then passed to the image generation, which uses GLIGEN \cite{gligen} and HumanSD \cite{humansd} to generate pose-aligned images.

\begin{itemize}
    \item \textbf{GLIGEN} extends text-to-image diffusion models by incorporating grounding inputs, including bounding boxes, object labels, and keypoints.
    In our experiments, we condition GLIGEN on keypoints derived from textual prompts, enabling it to generate images consistent with the described human poses.
    \item \textbf{HumanSD} fine-tunes the entire diffusion model using a heatmap-guided denoising loss that integrates pose information into the generation process.
    In our experiments, we condition HumanSD on skeleton images constructed from the extracted keypoints, enabling it to generate human images that accurately reflect the described poses.
\end{itemize}
\noindent The feedback system includes two modules: the first module evaluates the predicted keypoints using ChatGPT o1-preview \cite{OpenAI2024o1}, and the second evaluates the generated image using ChatGPT 4o \cite{openai_gpt4o}. This feedback loop iteratively refines the predicted keypoints and the generated image, improving their alignment with the input prompt.

\paragraph{Template for Keypoint Generator}
Our template consists of two parts: 1) keypoint generation specification and 2) base instruction.
The specification defines the task objective, describes the 17-point keypoint format, and provides guidelines for anatomical and positional constraints. It also includes a structured procedure for generating accurate 3D human keypoints from textual input. The base instruction explicitly directs the model to follow these definitions and produce outputs that conform to the expected structure. Our template is structured as follows:

{\scriptsize
\[
\begin{array}{l}
1) \left\{ \begin{array}{l}
\text{Objective: Generate 3D human keypoints.} \\
\text{...} \\
\text{Keypoint Format: [x0, y0, z0], Nose}\\
\text{...} \\
\text{Guidelines: ``Special points'' in contact with the ground must have $z = 0$. }\\
\text{...} \\
\text{Process steps: (1) Interpret the prompt and extract pose-specific features.}\\
\text{...} \\
\end{array} \right. \\[5mm]
2) \left\{ \begin{array}{l}
\text{Follow the steps closely and accurately identify} \\
\text{objects based on the given prompt.} \\
\text{Ensure adherence to the above output format.}\\
\end{array} \right.
\end{array}
\]
}
\paragraph{Template for Feedback System}
Similar to the keypoint generator, the template for the first feedback module consists of two parts: 1) keypoint generation specification and 2) base instruction. The specification is identical to that of the keypoint generator, defining the task, keypoint format, guidelines, and generation steps. However, the base instruction serves a different role: it makes the first feedback module to evaluate whether the generated keypoints accurately reflect the pose described in the prompt. If any misalignment is identified, the module detects incorrect aspects of the generated keypoints and regenerates them accordingly.

{\scriptsize
\[
\begin{array}{l}
1) \left\{ \begin{array}{l}
\text{Objective: Generate 3D human keypoints.} \\
\text{...} \\
\text{Keypoint Format: [x0, y0, z0], Nose}\\
\text{...} \\
\text{Guidelines: ``Special points'' in contact with the ground must have $z = 0$. }\\
\text{...} \\
\text{Process steps: (1) Interpret the prompt and extract pose-specific features.}\\
\text{...} \\
\end{array} \right. \\[5mm]
2) \left\{ \begin{array}{l}
\text{Determine whether the keypoints match the prompt. } \\
\text{If not, revise the incorrect analysis, regenerate the keypoints,} \\
\text{and explain the reason.}\\
\end{array} \right.
\end{array}
\]
}

\noindent The second feedback system consists of a single component: a base instruction that evaluates whether the generated image aligns with the input prompt.

\paragraph{Pose-related prompts} We describe the pose-related prompts used in our experiments. To evaluate the overall performance of our model, we employed 15 pose prompts in total: five each for common, acrobatic, and yoga categories. The general pose prompts include common activities such as standing and raising one hand, whereas the acrobatic and yoga prompts use the precise names of poses that are not typically seen in daily activities. The prompts were designed to include only essential pose-related details, thereby enabling focused evaluation of action depiction and demonstrating coverage of diverse poses. The prompts used in our experiments follow the format: 

\begin{center}
    \textit{``A person is performing \{pose\}.''}
\end{center}

The specific pose substituted into \{\textit{pose}\} will be described below.

\begin{itemize}
  \item Yoga poses
    \begin{itemize}
      \item The Cow and cat pose: A pair of poses performed on hands and knees, where Cow Pose arches the back and lifts the chest, while Cat Pose rounds the spine with the chin tucked.
      \item The Downward dog pose: An inverted V-shaped pose where both hands and feet touch the ground, with hips raised upward. It stretches the hamstrings, calves, and shoulders.
      \item The Boat pose: A core-strengthening pose where the person balances on their sitting bones with both legs and arms extended forward, forming a "V" shape.
      \item The Split pose: A flexibility pose where the legs are extended in opposite directions along the floor (front and back), emphasizing hip and hamstring mobility.
      \item The Tree pose: A standing balance pose where one foot is placed on the inner thigh or calf of the opposite leg, and hands are brought together overhead or at the chest.
    \end{itemize}
\end{itemize}

\begin{itemize}
  \item Acrobatic poses
    \begin{itemize}
      \item Hollow hold: A core-focused pose where the person lies on their back, lifting both legs and shoulders off the ground to form a curved, hollow body shape.
      \item L-sit: An isometric strength pose where the body is supported on straight arms, with both legs extended forward to form an “L” shape.
      \item One-arm handstand: A vertical balancing pose where the entire body is supported by a single arm, requiring high strength and stability.
      \item Windmill: A twisting pose with legs apart, where one arm reaches down and the other up, often used to train flexibility and coordination.
      \item Y-scale: A standing balance pose where one leg is lifted high to the side, forming a “Y” shape with the body, showing flexibility and control.
    \end{itemize}
\end{itemize}

\begin{itemize}
  \item Common poses: Self-explanatory poses
    \begin{itemize}
      \item Kicking
      \item Raising both hands
      \item Raising one hand
      \item Punching
      \item Standing
    \end{itemize}
\end{itemize}

\paragraph{Baselines} We compare our framework with SD 1.4 \cite{sd} and SDXL \cite{sdxl}, two representative and high-performing text-to-image (T2I) models. These baselines were selected because our image generators---GLIGEN and HumanSD---are both built upon the SD architecture.

\paragraph{Metrics} The effectiveness of our framework is demonstrated by evaluating the final output images. To assess text-image alignment, we utilize CLIPScore \cite{hessel2021clipscore} and VQAScore \cite{lin2024evaluating}. CLIPScore measures alignment by computing the cosine similarity between text and image embeddings obtained from the CLIP model \cite{radford2021learning}. While CLIPScore is effective for capturing overall semantic correspondence, it has difficulty handling compositional text prompts involving actions \cite{lin2024evaluating}. To address this limitation, we additionally employ VQAScore, which evaluates the accurate depiction of specific human actions. By using generative vision-language models, VQAScore performs compositional reasoning through answer generation, providing a more detailed assessment of action fidelity \cite{lin2024evaluating}. Additionally, for yoga poses, we employ ResNet \cite{he2016deep} classifier fine-tuned on the Yoga-82 dataset \cite{verma2020yoga}, which achieves 97.94\% accuracy, to demonstrate the discriminability among images generated for different yoga actions.

\begin{table*}[ht]
    \centering
    \begin{subtable}
    \centering
    
    \scalebox{0.5}{
        \begin{tabular}{ccccccccccccccccccccc}
        \toprule
      Yoga Pose & \multicolumn{2}{c}{Cow and cat pose} & \multicolumn{2}{c}{Downward dog pose} & \multicolumn{2}{c}{Boat pose} & \multicolumn{2}{c}{Split pose} & \multicolumn{2}{c}{Tree pose} & \multicolumn{2}{c}{AVG} \\ 
      \cmidrule(lr){2-3}
        \cmidrule(lr){4-5}
        \cmidrule(lr){6-7}
        \cmidrule(lr){8-9}
        \cmidrule(lr){10-11}
        \cmidrule(lr){12-13}
    Scores  & \multicolumn{1}{c}{VQA $\uparrow$} & \multicolumn{1}{c}{CLIP $\uparrow$} & \multicolumn{1}{c}{VQA $\uparrow$} & \multicolumn{1}{c}{CLIP $\uparrow$} & \multicolumn{1}{c}{VQA $\uparrow$} & \multicolumn{1}{c}{CLIP $\uparrow$} & \multicolumn{1}{c}{VQA $\uparrow$} 
    & \multicolumn{1}{c}{CLIP $\uparrow$} & \multicolumn{1}{c}{VQA $\uparrow$}& \multicolumn{1}{c}{CLIP $\uparrow$} & \multicolumn{1}{c}{VQA $\uparrow$} & \multicolumn{1}{c}{CLIP $\uparrow$}\\
      \midrule
      SD 1.4 & 0.097 & 0.235 & 0.153 & 0.240 & 0.001 & \textbf{0.236} & 0.133 & 0.228 & 0.055 & 0.245 & 0.088 & 0.237\\
      SDXL & 0.079 & 0.243 & 0.298 & 0.244 & 0.001 & 0.222 & 0.279 & 0.233 & 0.438 & \textbf{0.256} & 0.219 & 0.240 \\
      \midrule
      Ours (HumanSD) & \underline{0.714} & \underline{0.244} & \underline{0.924} & \textbf{0.264} & \textbf{0.816} & \underline{0.231} & \textbf{1.000} & \textbf{0.259} & \underline{0.765} & \underline{0.249} & \underline{0.844} & \textbf{0.249} \\
      Ours (GLIGEN) & \textbf{0.926} & \textbf{0.250} & \textbf{0.998} & \underline{0.262} & \underline{0.669} & 0.225 & \underline{0.999} & \underline{0.251} & \textbf{0.904} & 0.248 & \textbf{0.899} & \underline{0.247}\\      
      \midrule \midrule
      Yoga82 & 0.905 & 0.247 & 0.933 & 0.253 & 0.835 & 0.266 & 0.916 & 0.259 & 0.913 & 0.256 & 0.900 & 0.256\\
      \bottomrule 
    \end{tabular}
    }
    \caption{\textbf{Quantitative comparison for yoga poses. }}
    \label{tab:yoga-score}
    \vspace{-0.5cm}
    \end{subtable}

    
    \begin{subtable}
    \centering
    
    \scalebox{0.5}{
    \begin{tabular}{ccccccccccccccccccccc}
      \toprule
      Acrobatic Pose & \multicolumn{2}{c}{Hollow hold pose} & \multicolumn{2}{c}{L-sit pose} & \multicolumn{2}{c}{One-arm handstand pose} & \multicolumn{2}{c}{Windmill pose} & \multicolumn{2}{c}{Y-scale pose} & \multicolumn{2}{c}{AVG} \\ 
      \cmidrule(lr){2-3}
        \cmidrule(lr){4-5}
        \cmidrule(lr){6-7}
        \cmidrule(lr){8-9}
        \cmidrule(lr){10-11}
        \cmidrule(lr){12-13}
    Scores  & \multicolumn{1}{c}{VQA $\uparrow$} & \multicolumn{1}{c}{CLIP $\uparrow$} & \multicolumn{1}{c}{VQA $\uparrow$} & \multicolumn{1}{c}{CLIP $\uparrow$} & \multicolumn{1}{c}{VQA $\uparrow$} & \multicolumn{1}{c}{CLIP $\uparrow$} & \multicolumn{1}{c}{VQA $\uparrow$} 
    & \multicolumn{1}{c}{CLIP $\uparrow$} & \multicolumn{1}{c}{VQA $\uparrow$}& \multicolumn{1}{c}{CLIP $\uparrow$} & \multicolumn{1}{c}{VQA $\uparrow$} & \multicolumn{1}{c}{CLIP $\uparrow$}\\
      \midrule
      SD 1.4 & 0.001 & 0.191 & 0.001 & 0.208 & 0.081 & 0.245 & 0.011 & 0.231 & 0.001 & 0.197 & 0.019 & 0.215\\
      SDXL & 0.001 & 0.191 & 0.001 & \textbf{0.228} & 0.001 & 0.243 & 0.004 & 0.233 & 0.054 & 0.218 & 0.013 & 0.223 \\
      \midrule
      Ours (HumanSD) & \underline{0.975} & \underline{0.205} & \underline{0.757} & \underline{0.224} & \textbf{0.997} & \textbf{0.267} & \underline{0.803} & \textbf{0.242} & \textbf{0.999} & \textbf{0.241} & \underline{0.907} & \textbf{0.236}  \\
      Ours (GLIGEN) & \textbf{0.999} & \textbf{0.211} & \textbf{0.807} & 0.216 & \underline{0.831} & \underline{0.261} & \textbf{0.960} & \underline{0.241} & \underline{0.989} & \underline{0.226} & \textbf{0.918} & \underline{0.231} \\ 
    
      \bottomrule
    \end{tabular}
    }
    \caption{\textbf{Quantitative comparison for acrobatic poses.}}
    \label{tab:acrobatic-score}
    \vspace{-0.5cm}
    \end{subtable}

    
    \begin{subtable}
    \centering

        \scalebox{0.5}{
    \begin{tabular}{ccccccccccccccccccccc}
      \toprule
      Common Pose  & \multicolumn{2}{c}{Kicking} & \multicolumn{2}{c}{Raising both hands} & \multicolumn{2}{c}{Raising one hand} & \multicolumn{2}{c}{Punching} & \multicolumn{2}{c}{Standing} & \multicolumn{2}{c}{AVG} \\ 
      \cmidrule(lr){2-3}
        \cmidrule(lr){4-5}
        \cmidrule(lr){6-7}
        \cmidrule(lr){8-9}
        \cmidrule(lr){10-11}
        \cmidrule(lr){12-13}
    
     Scores & \multicolumn{1}{c}{VQA $\uparrow$} & \multicolumn{1}{c}{CLIP $\uparrow$} & \multicolumn{1}{c}{VQA $\uparrow$} & \multicolumn{1}{c}{CLIP $\uparrow$} & \multicolumn{1}{c}{VQA $\uparrow$}
    & \multicolumn{1}{c}{CLIP $\uparrow$} & \multicolumn{1}{c}{VQA $\uparrow$}& \multicolumn{1}{c}{CLIP $\uparrow$} & \multicolumn{1}{c}{VQA $\uparrow$}& \multicolumn{1}{c}{CLIP $\uparrow$} & \multicolumn{1}{c}{VQA $\uparrow$}& \multicolumn{1}{c}{CLIP $\uparrow$}\\
      \midrule
      SD 1.4 & 0.912 & 0.251 & 0.778 & 0.221 & 0.629 & 0.217 & 0.806 & 0.230 & \underline{1.000} & 0.216 & 0.825  & 0.227 \\
      SDXL & 0.933 & 0.251 & 0.768 & 0.226 & 0.633 & 0.210 & 0.831 & 0.227 & \textbf{1.000} & \underline{0.234}  & 0.833 & 0.229 \\
      \midrule
      Ours (HumanSD) & \textbf{1.000} & \textbf{0.259} & \textbf{1.000} & \textbf{0.230} & \underline{0.838}  & \textbf{0.236} & \textbf{0.939} & \textbf{0.239} & 1.000 & \textbf{0.253} & \underline{0.956} & \textbf{0.242} \\      
      Ours (GLIGEN) & \underline{0.992} & \underline{0.252} & \underline{0.996} & \underline{0.227} & \textbf{0.999} & \underline{0.232} & \underline{0.913} & \underline{0.231} & 1.000 & 0.207 & \textbf{0.980} & \underline{0.230} \\   
      \bottomrule
    \end{tabular}
    }
    \caption{\textbf{Quantitative comparison for common poses.}}
    \label{tab:common-score}
    \end{subtable}

    \captionsetup{labelformat=empty}
    \caption{Table 1-3: \textbf{Quantitative comparison of image-prompt similarity.} Image-prompt similarity is evaluated using VQAScore (VQA) and CLIPScore (CLIP). \textbf{Bold} : Best, \underline{underline} : second best.
    Yoga poses are also compared with Yoga-82 dataset.}
    \addtocounter{table}{-1}
    \label{tab:scores-poses}

\end{table*}

\begin{table}[ht]
    \centering
    
        \scalebox{0.65}{
    \begin{tabular}{ccccccccccccccccccccc}
      \toprule
      Yoga Pose & Cow and cat pose & Downward dog pose & Boat pose & Split pose & Tree pose & AVG \\ 

      \midrule
      SD 1.4 & 31.3\% & 37.5\% & 6.2\% & 50.0\% & 75.0\% & 40.0\%\\
      SDXL & 68.7 \% & 81.8 \% & 18.8\% & 68.8 \% & 87.5\% & 65.1\% \\
      \midrule
      Ours (HumanSD) & \textbf{100\%} & \underline{93.3\%} & \underline{80\%} & \underline{96.1\%} & \textbf{100.0\%} & \underline{93.9\%} \\
      Ours (GLIGEN) & \textbf{100\%} & \textbf{96.7\%} & \textbf{87.9\%}& \textbf{100.0\%} & \textbf{100.0\%} & \textbf{96.9\%} \\

      \bottomrule
    \end{tabular}
    }
    \vspace{10 pt}
    \caption{ \textbf{Quantitative comparison of classification accuracy.} The ResNet model, fine-tuned on the Yoga-82 dataset \cite{verma2020yoga}, is used to assess the generated yoga poses. \textbf{Bold} : Best, \underline{underline} : second best.
    }
    \label{tab:yoga-classification}
    
\end{table}

\subsection{Pose-Aware Text-to-Image Alignment}

\paragraph{Quantitative Evaluation}
 \Cref{tab:yoga-score,tab:acrobatic-score,tab:common-score} present the results of VQAScore \cite{lin2024evaluating} and CLIPScore \cite{hessel2021clipscore} for yoga, acrobatic, and common poses, respectively. In addition, we report classification accuracy using a ResNet model fine-tuned on the Yoga-82 dataset \cite{verma2020yoga} to further evaluate the yoga poses, as shown in \Cref{tab:yoga-classification}. 

In \Cref{tab:yoga-score,tab:acrobatic-score}, our framework achieves superior VQAScore and CLIPScore despite the complexity of yoga and acrobatic poses. As VQAScore is more sensitive to pose-specific features than CLIPScore \cite{lin2024evaluating}, significantly lower VQAScores of the baselines highlight their difficulty in modeling detailed poses. In contrast, our framework maintains superior prompt-image alignment. Moreover, \Cref{tab:common-score} shows that our model performs well on simple pose prompts, achieving higher VQAScores compared to the baselines. This improvement suggests that our framework enables more accurate image-text alignment, even for typical input prompts. Overall, these results demonstrate that PointT2I consistently outperforms the baselines across most pose categories, particularly in VQAScore, which better reflects fine-grained prompt understanding. This indicates that our framework more precisely interprets and renders the pose descriptions embedded in the prompts.

 Furthermore, we evaluate how well the generated images represent yoga poses, using a ResNet classifier trained on the Yoga-82 dataset \cite{verma2020yoga}. As demonstrated in \Cref{tab:yoga-classification}, the images generated by our framework achieve significantly higher classification accuracy. In particular, for the boat pose, which SD and SDXL have difficulty rendering accurately, PointT2I shows marked improvements, as evidenced by \Cref{tab:yoga-score} and \Cref{tab:yoga-classification}.
Taken together, \Cref{tab:yoga-score,tab:acrobatic-score,tab:common-score,tab:yoga-classification} show that our proposed framework significantly outperforms the baselines in all evaluation metrics. The relative improvement is especially pronounced in challenging cases such as yoga and acrobatic poses, as reflected by the significantly higher VQAScore achieved by our method. This result highlights that our framework effectively captures fine-grained pose semantics from the prompts.

\begin{figure}[ht]
    \centering
    \includegraphics[width=0.9\linewidth]{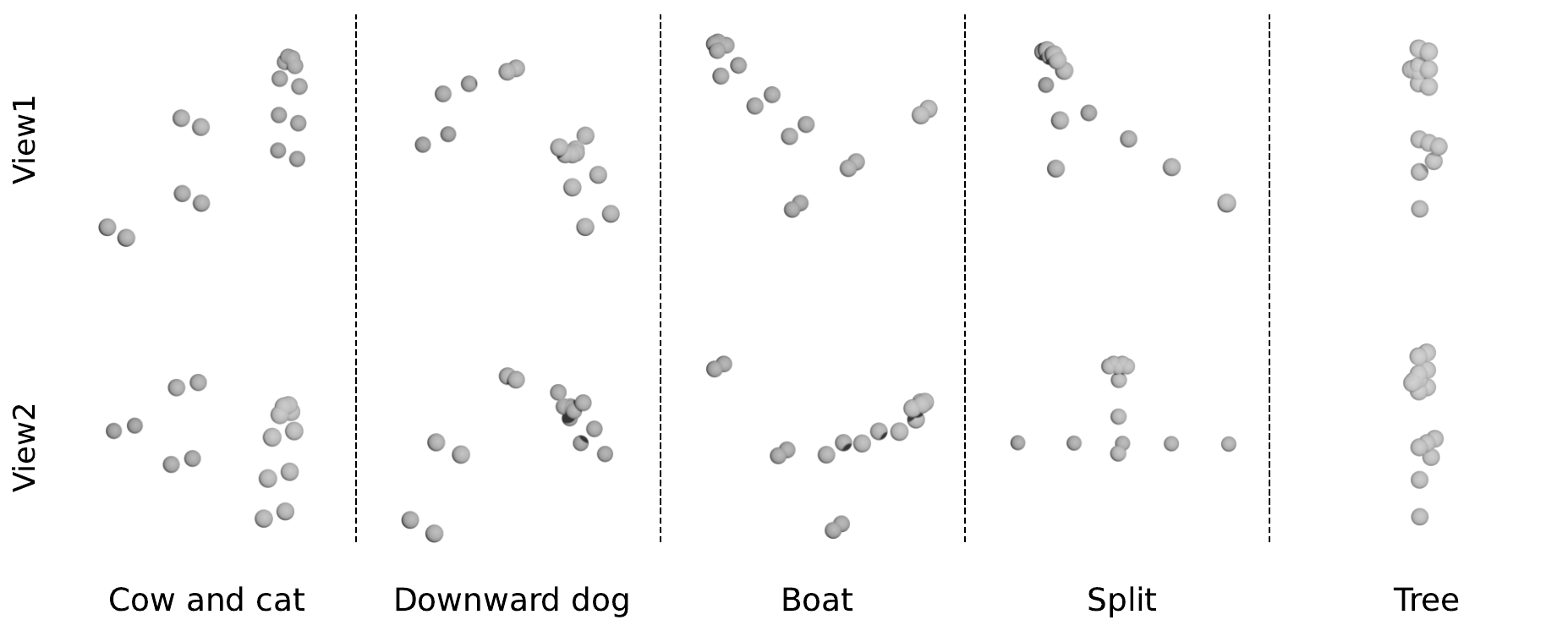}
    \caption{\textbf{Generated keypoints for yoga pose prompts.}}
    \label{fig:keypoints_yoga}
\end{figure}

\begin{figure}[ht]
    \centering
    \includegraphics[width=0.9\linewidth]{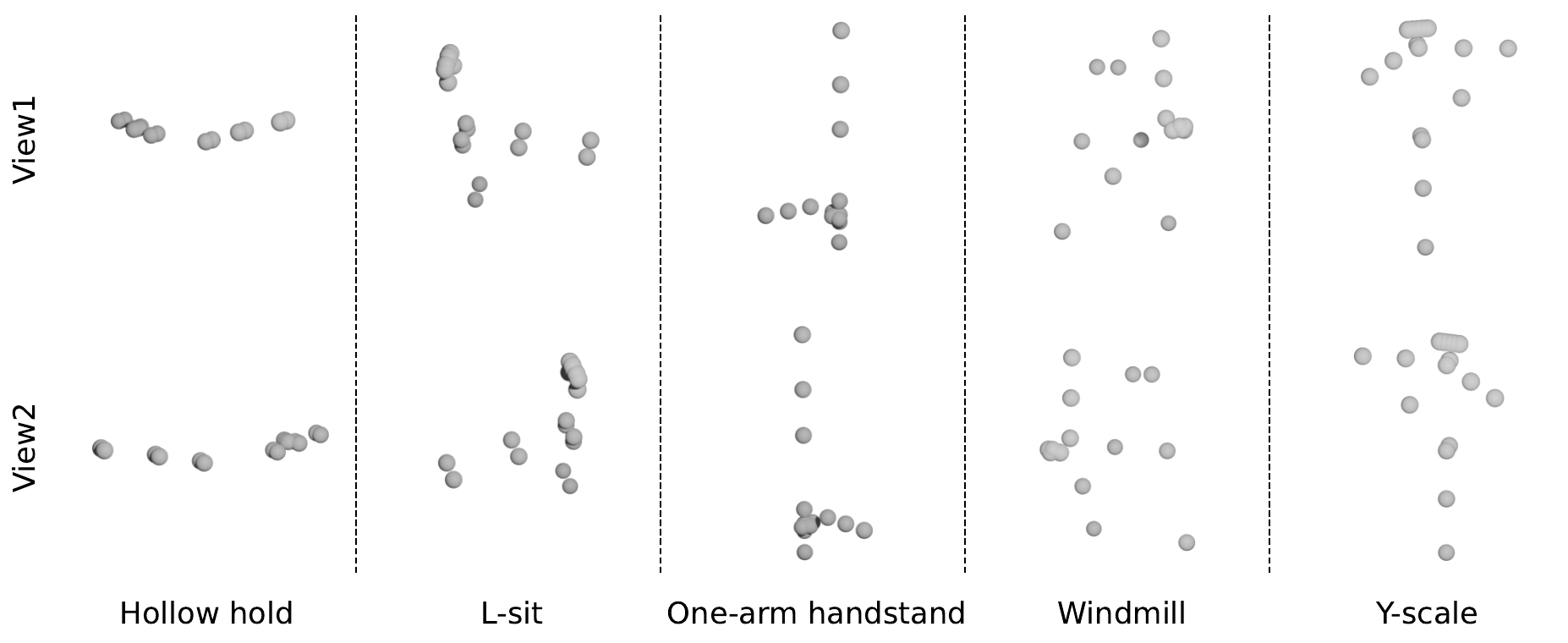}
    \caption{ \textbf{Generated keypoints for acrobatic pose prompts.} }
    \label{fig:keypoints_acrobatic}
\end{figure}

\begin{figure}[ht]
    \centering
    \includegraphics[width=0.9\linewidth]{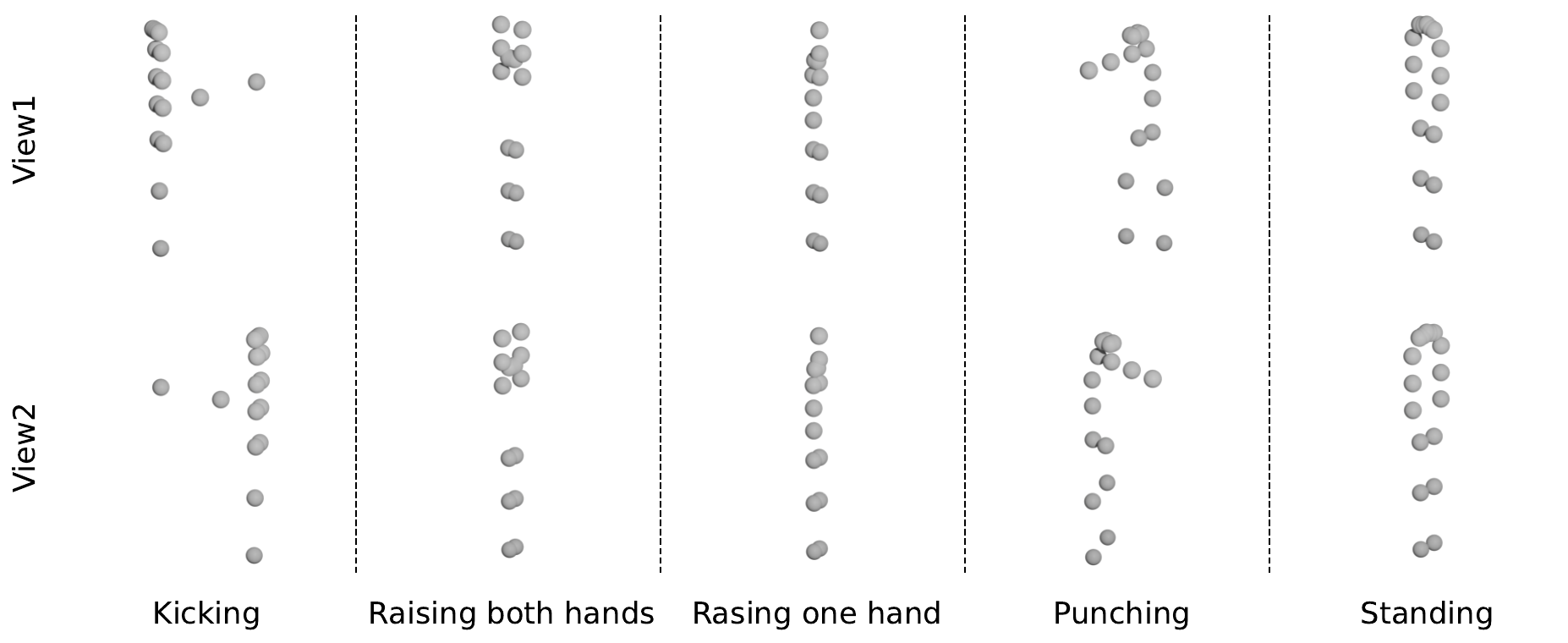}
    \caption{ \textbf{Generated keypoints for common pose prompts.} }
    \label{fig:keypoints_common}
\end{figure}

\paragraph{Qualitative Evaluation}
Now, we present qualitative results for 3D keypoints and the output images to provide a more comprehensive assessment of our framework's performance. 
\Cref{fig:keypoints_yoga,fig:keypoints_acrobatic,fig:keypoints_common} illustrate keypoints generated for various poses from multiple viewpoints. We observe that the output keypoints accurately capture the important characteristics of each pose. 
For instance, in \Cref{fig:keypoints_yoga}, the keypoints for the split pose clearly reflect fully extended legs in opposite directions, while in \Cref{fig:keypoints_acrobatic}, the keypoints for the y-scale pose depict a vertically lifted leg that forms a distinctive Y-shape. These results show that the keypoint generator can accurately generate 3D pose keypoints aligned with the input prompt.

\clearpage 

\begin{figure*}[ht] 
    \centering
        \includegraphics[width=1\textwidth]{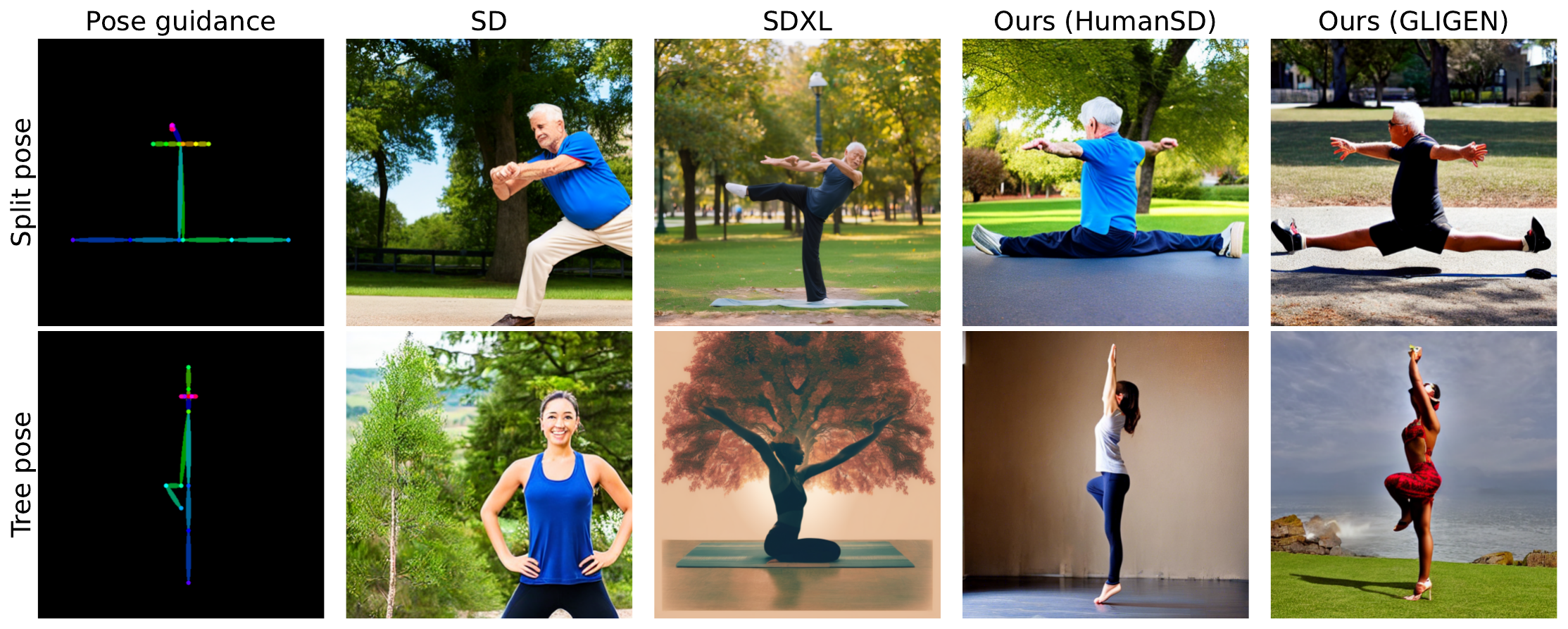}
    \caption{\textbf{Qualitative comparison with the baselines on yoga poses.} The baselines fail to capture correct body structure, while our method preserves pose fidelity via keypoint-based guidance.
    }
    \label{fig:qualitative-2dimage}
    \vspace{-8pt}
\end{figure*}

\begin{figure}[ht]
    \centering
    \vspace{2.5cm}
    \hspace*{-1.7cm}
    \includegraphics[width=1\linewidth]{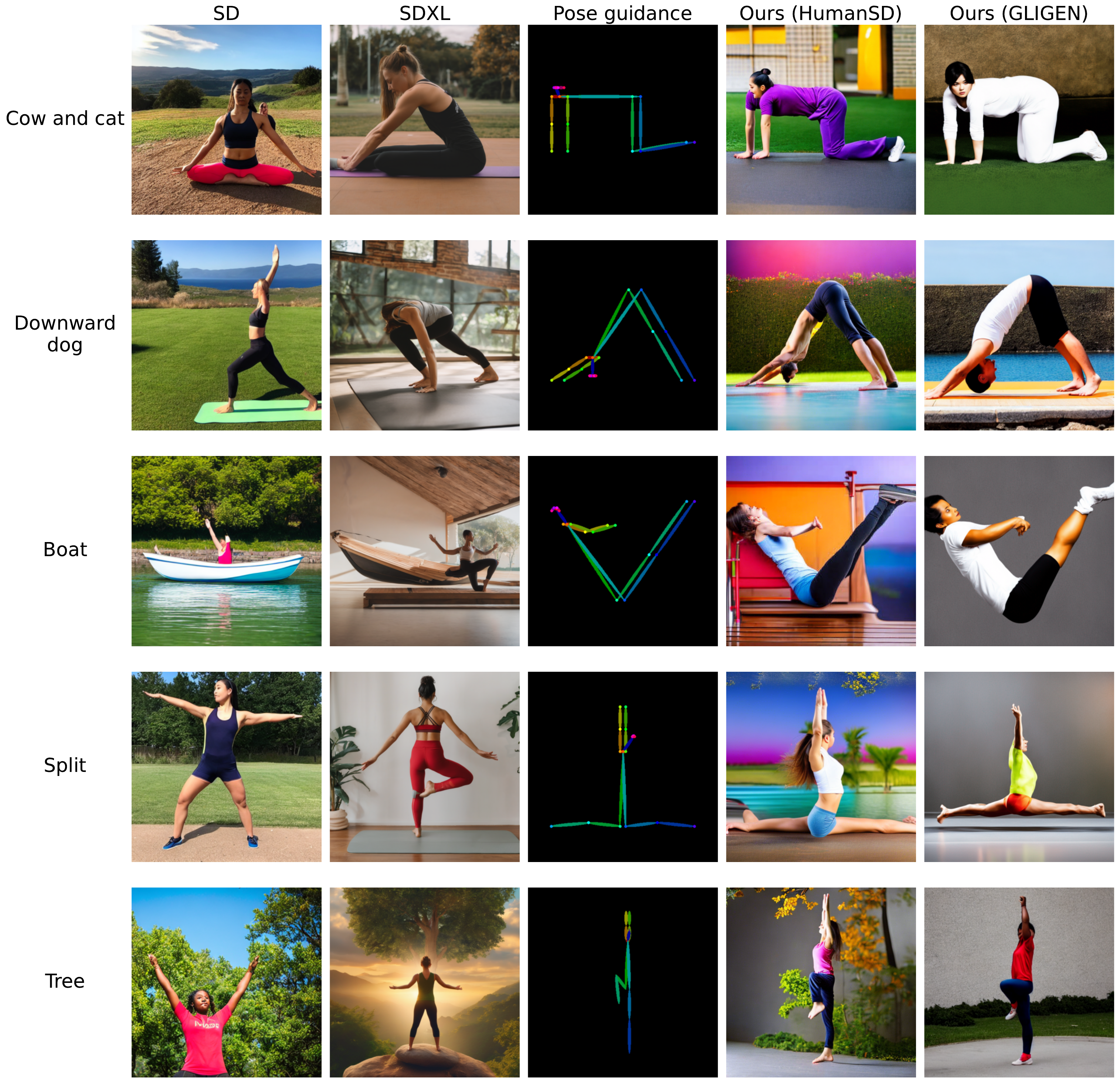}
    \caption{\textbf{Generated images from yoga pose prompts.}}
    \label{fig:samples-yoga}
\end{figure}

\begin{figure}[ht]
    \centering
    \vspace{2.5cm}
    \hspace*{-1.7cm}
    \includegraphics[width=1\linewidth]{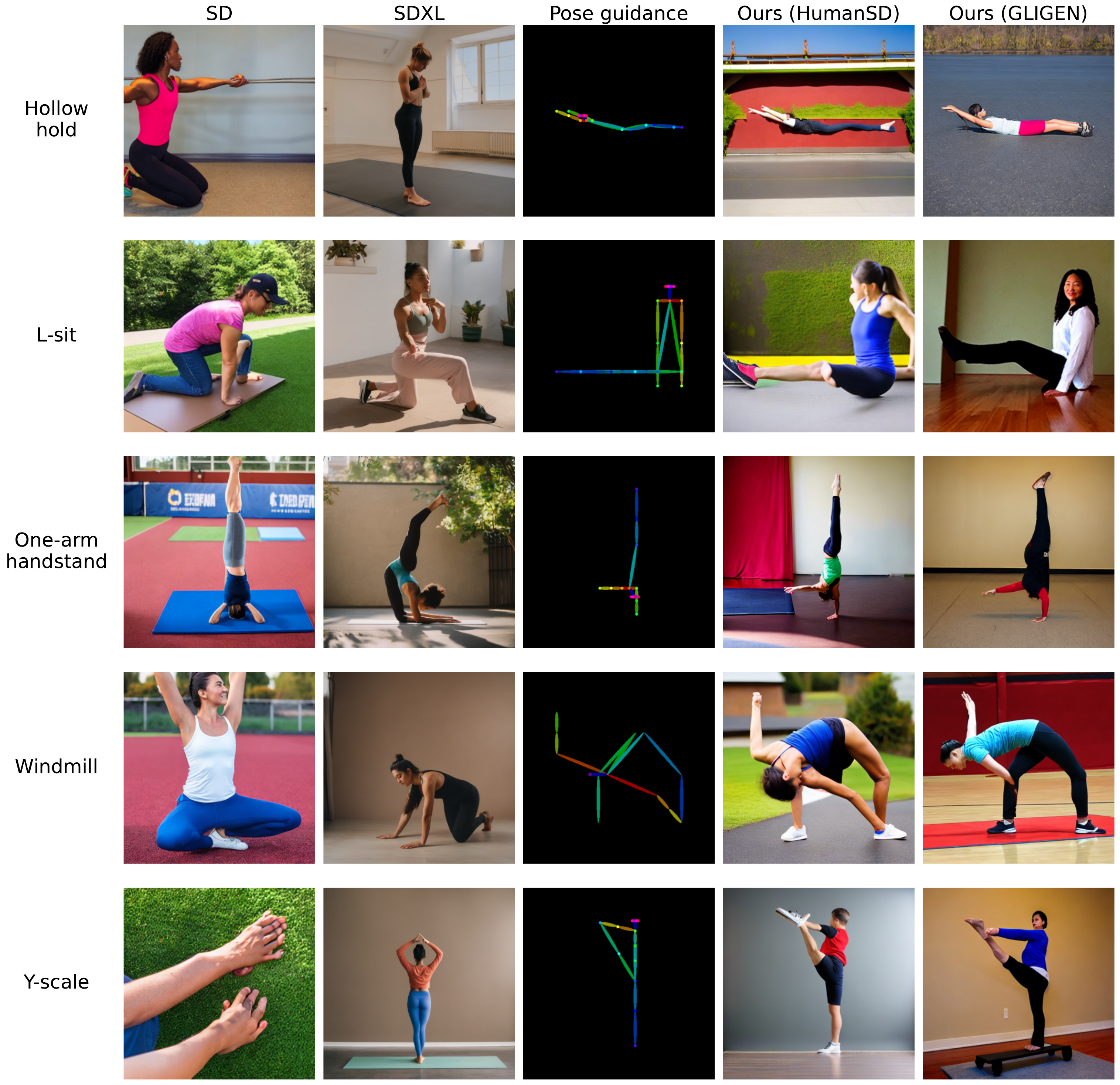}
    \caption{\textbf{Generated images from acrobatic pose prompts.}}
    \label{fig:samples-acrobatic}
\end{figure}

\begin{figure}[ht]
    \centering
    \vspace{2.5cm}
    \hspace*{-1.7cm}
    \includegraphics[width=1\linewidth]{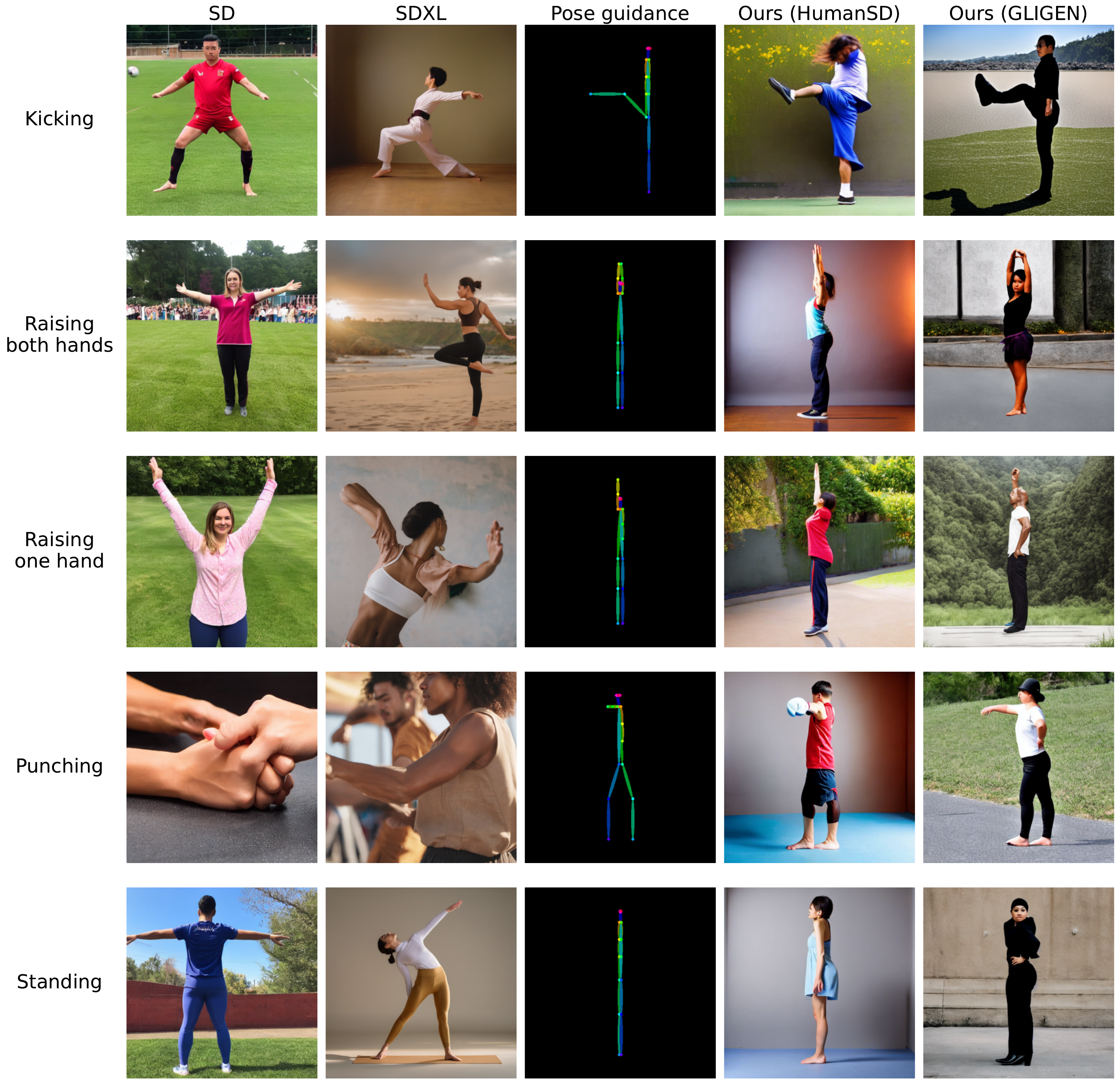}
    \caption{\textbf{Generated images from common pose prompts.}}
    \label{fig:samples-common}
\end{figure}

\Cref{fig:qualitative-2dimage} presents a qualitative comparison of generated images for two challenging poses: split (top row) and tree (bottom row). The leftmost column shows the pose guidance, obtained by projecting the 3D keypoints inferred from textual prompts. This guidance is then used as a conditioning input for image generation.
The baselines, SD and SDXL, fail to reproduce the correct body configurations, highlighting the limitations of prompt-only guidance in modeling fine-grained human poses. In particular, both models miss critical structural details---such as the horizontal leg extension in the split pose and the vertical leg lift in the tree pose.
In contrast, when guided by our LLM-predicted pose keypoints, the underlying rendering capabilities of pose-aware models---such as HumanSD and GLIGEN---are effectively activated to reproduce the target poses described in the prompts accurately. These models, while unable to interpret pose semantics directly from language, can faithfully render complex human configurations once provided with explicit structural guidance inferred from text. By generating such pose guidance directly from natural language, our framework maximizes the utility of these pose-aware backbones, enabling them to perform beyond their original limitations. \Cref{fig:samples-yoga,fig:samples-acrobatic,fig:samples-common} provide additional qualitative comparisons across a broader range of human poses. Notably, for poses such as the boat pose and the tree pose in \Cref{fig:samples-yoga}, the baselines often misinterpret the prompts and generate images that include a boat or a tree, rather than depicting the corresponding yoga posture. In contrast, our pose-guided approach produces images that accurately capture the target pose. 

These results demonstrate that our framework does more than supply pose guidance---it unlocks the latent structural rendering potential of existing T2I backbones by bridging the semantic gap between natural language and pose specification. By translating textual intent into pose structures, PointT2I enables accurate, pose-conditioned image generation without reliance on external pose annotations or fine-tuning.

\newpage

\newpage 

\clearpage

\section{Ablation Study}
\label{sec:ablation}

In this section, we conduct ablation studies to assess the effectiveness and versatility of our framework. We examine the contribution of the LLM-based feedback system in enhancing pose fidelity through multi-round feedback. We also assess the robustness of our approach under diverse prompt formulations. Additionally, we evaluate the compatibility of our method with multiple image generators.

\subsection{Necessity of Feedback System}
\label{sec:feedback}
One of the main contributions of PointT2I is the LLM-based feedback system, which ensures that the generated output faithfully captures the actions intended by the prompt. To assess the utility of the feedback system, we separately evaluate each of its two modules by varying the number of feedback rounds for one, while keeping the other at three rounds. For this experiment, the 5 acrobatic pose prompts are provided as input, and GLIGEN is used to generate the corresponding image.

\begin{figure*}[ht] 
    \centering
        \includegraphics[width=1\textwidth]{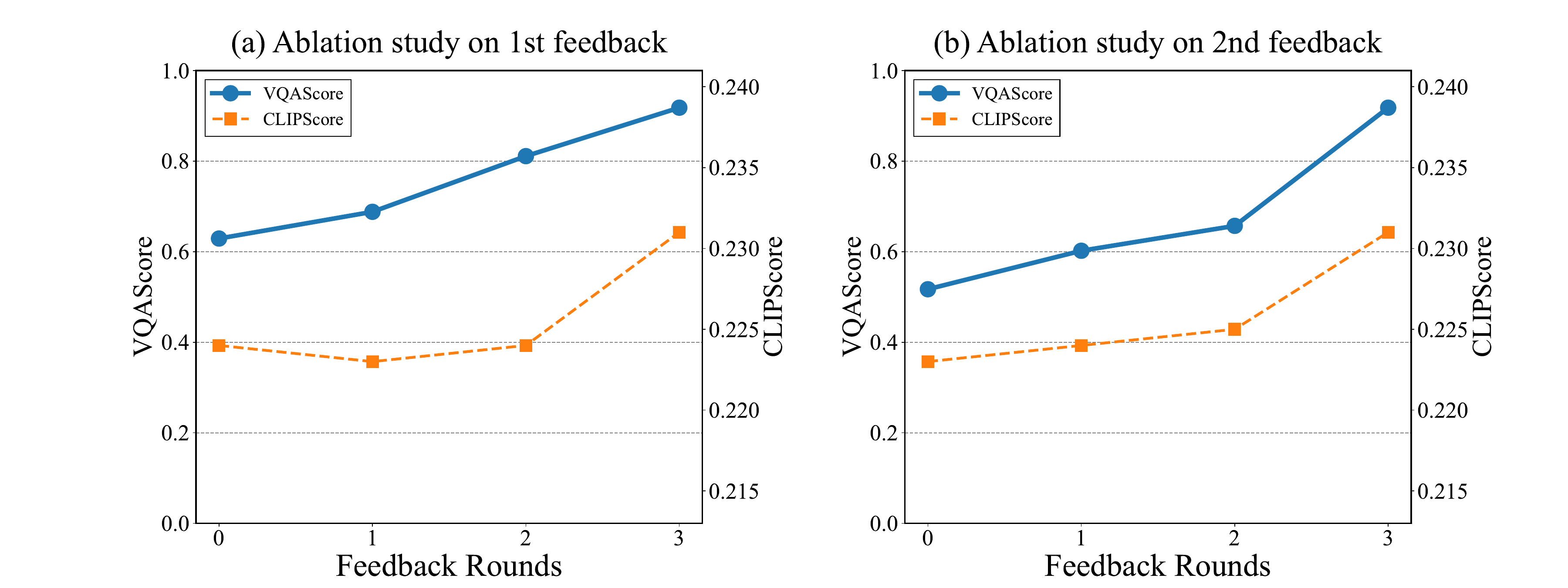}
    \caption{ \textbf{Necessity of the feedback system}. For each feedback module, increasing feedback rounds improves the text-image alignment performance.}
        
    \label{fig:ablation_feedback}
    \vspace{-8pt}
\end{figure*}

In \Cref{fig:ablation_feedback}, we plot the average VQAScore and CLIPScore evaluated over diverse poses, with respect to the number of feedback rounds. The figure demonstrates that both metrics improve consistently as feedback iterations increase across the two modules.
The improvement is particularly pronounced in VQAScore, which is more sensitive to human pose-specific details. To provide a more detailed result, \Cref{tab:ablation-1st-feedback,tab:ablation-2nd-feedback} report performance metrics for each individual pose. The results indicate that both feedback modules tend to improve performance under various pose conditions. These results validate the effectiveness of our LLM-based feedback system and underscore its significant contribution to the overall performance of the framework.

\begin{table*}[ht]
    \centering
    \scalebox{0.55}{
    \begin{tabular}{ccccccccccccccccccccc}
      \toprule
      Acrobatic pose  & \multicolumn{2}{c}{Hollow hold} & \multicolumn{2}{c}{L-sit} & \multicolumn{2}{c}{One-arm handstand} & \multicolumn{2}{c}{Windmill} & \multicolumn{2}{c}{Y-scale} & \multicolumn{2}{c}{AVG} \\ 
      \cmidrule(lr){2-3}
        \cmidrule(lr){4-5}
        \cmidrule(lr){6-7}
        \cmidrule(lr){8-9}
        \cmidrule(lr){10-11}
        \cmidrule(lr){12-13}
    
feedback rounds & \multicolumn{1}{c}{VQA\ $\uparrow$} & \multicolumn{1}{c}{CLIP\ $\uparrow$} & \multicolumn{1}{c}{VQA\ $\uparrow$} & \multicolumn{1}{c}{CLIP\ $\uparrow$} & \multicolumn{1}{c}{VQA\ $\uparrow$}
& \multicolumn{1}{c}{CLIP\ $\uparrow$} & \multicolumn{1}{c}{VQA\ $\uparrow$} & \multicolumn{1}{c}{CLIP\ $\uparrow$} & \multicolumn{1}{c}{VQA\ $\uparrow$} & \multicolumn{1}{c}{CLIP\ $\uparrow$} & \multicolumn{1}{c}{VQA\ $\uparrow$} & \multicolumn{1}{c}{CLIP\ $\uparrow$}\\
      \midrule
      0 & 0.829 & 0.183 & 0.278 & \underline{0.214} & 0.656 & 0.257 & \underline{0.829} & \underline{0.239} & 0.551 & \textbf{0.227} & 0.629 & \underline{0.224} \\
      1 & \underline{0.954} & 0.188 & 0.656 & 0.207 & \textbf{0.892} & \underline{0.260} & 0.706 & 0.236 & 0.231 & 0.221 & 0.688 & 0.223 \\
      2 & 0.933 & \underline{0.189} & \textbf{0.889 }& 0.214 & \underline{0.863} & 0.258 & 0.750  & 0.234 & \underline{0.618} & 0.224 & \underline{0.811} & 0.224 \\
      3 & \textbf{0.999} & \textbf{0.211} & \underline{0.807} & \textbf{0.216} & 0.831 & \textbf{0.261} & \textbf{0.960}  & \textbf{0.241} & \textbf{0.989} & \underline{0.226} & \textbf{0.918} & \textbf{0.231} \\
      \bottomrule
    \end{tabular}
    }
    \caption{ \textbf{Effectiveness of the first feedback module.} \textbf{Bold} : Best, \underline{underline} : second best.}
    \label{tab:ablation-1st-feedback}
\end{table*}

\begin{table*}[ht]
    \centering
    \scalebox{0.55}{
    \begin{tabular}{ccccccccccccccccccccc}
      \toprule
      Acrobatic pose  & \multicolumn{2}{c}{Hollow hold} & \multicolumn{2}{c}{L-sit} & \multicolumn{2}{c}{One-arm handstand} & \multicolumn{2}{c}{Windmill} & \multicolumn{2}{c}{Y-scale} & \multicolumn{2}{c}{AVG} \\ 
      \cmidrule(lr){2-3}
        \cmidrule(lr){4-5}
        \cmidrule(lr){6-7}
        \cmidrule(lr){8-9}
        \cmidrule(lr){10-11}
        \cmidrule(lr){12-13}
    
feedback rounds & \multicolumn{1}{c}{VQA\ $\uparrow$} & \multicolumn{1}{c}{CLIP\ $\uparrow$} & \multicolumn{1}{c}{VQA\ $\uparrow$} & \multicolumn{1}{c}{CLIP\ $\uparrow$} & \multicolumn{1}{c}{VQA\ $\uparrow$}
& \multicolumn{1}{c}{CLIP\ $\uparrow$} & \multicolumn{1}{c}{VQA\ $\uparrow$} & \multicolumn{1}{c}{CLIP\ $\uparrow$} & \multicolumn{1}{c}{VQA\ $\uparrow$} & \multicolumn{1}{c}{CLIP\ $\uparrow$} & \multicolumn{1}{c}{VQA\ $\uparrow$} & \multicolumn{1}{c}{CLIP\ $\uparrow$}\\

      \midrule
      0 & 0.739 & 0.187 & 0.492 & \underline{0.214} & 0.654 & 0.258 & 0.692 & 0.237 & 0.007 & 0.219 & 0.517 & 0.223 \\
      1 & 0.815 & 0.187 & \underline{0.647} & 0.213 & 0.732 & \underline{0.261} & \underline{0.715} & 0.238 & 0.099 & 0.219 & 0.602 & 0.225 \\
      2 & \underline{0.992} & \underline{0.190} & 0.643 & 0.211 & \underline{0.778} & \textbf{0.262} & 0.679 & \underline{0.240} & \underline{0.193} & \underline{0.221} & \underline{0.657} & \underline{0.225} \\
      3 & \textbf{0.999} & \textbf{0.211} & \textbf{0.807} & \textbf{0.216} & \textbf{0.831} & 0.261 & \textbf{0.960} & \textbf{0.241} & \textbf{0.989} & \textbf{0.226} & \textbf{0.918} & \textbf{0.231} \\
      \bottomrule
    \end{tabular}
    }
    \caption{ \textbf{Effectiveness of the second feedback module.} \textbf{Bold} : Best, \underline{underline} : second best.}
    \label{tab:ablation-2nd-feedback}
\end{table*}

\begin{figure}[ht]
    \centering
    \includegraphics[width=1\linewidth]{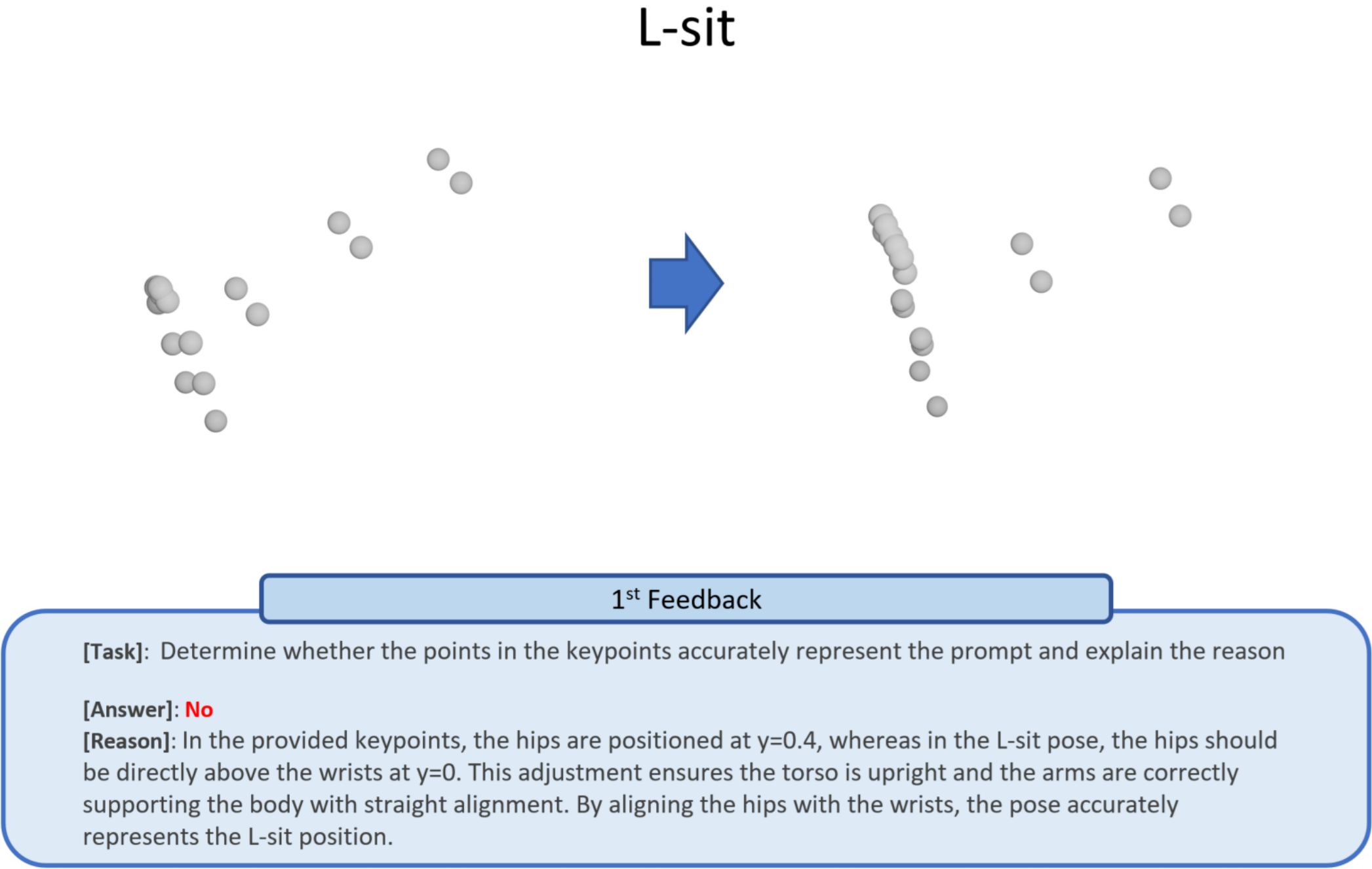}
    \caption{ \textbf{First feedback module for the L-sit pose.}}
    \label{fig:feedback-sample-yoga}
\end{figure}

\begin{figure}[ht]
    \centering
    \includegraphics[width=1\linewidth]{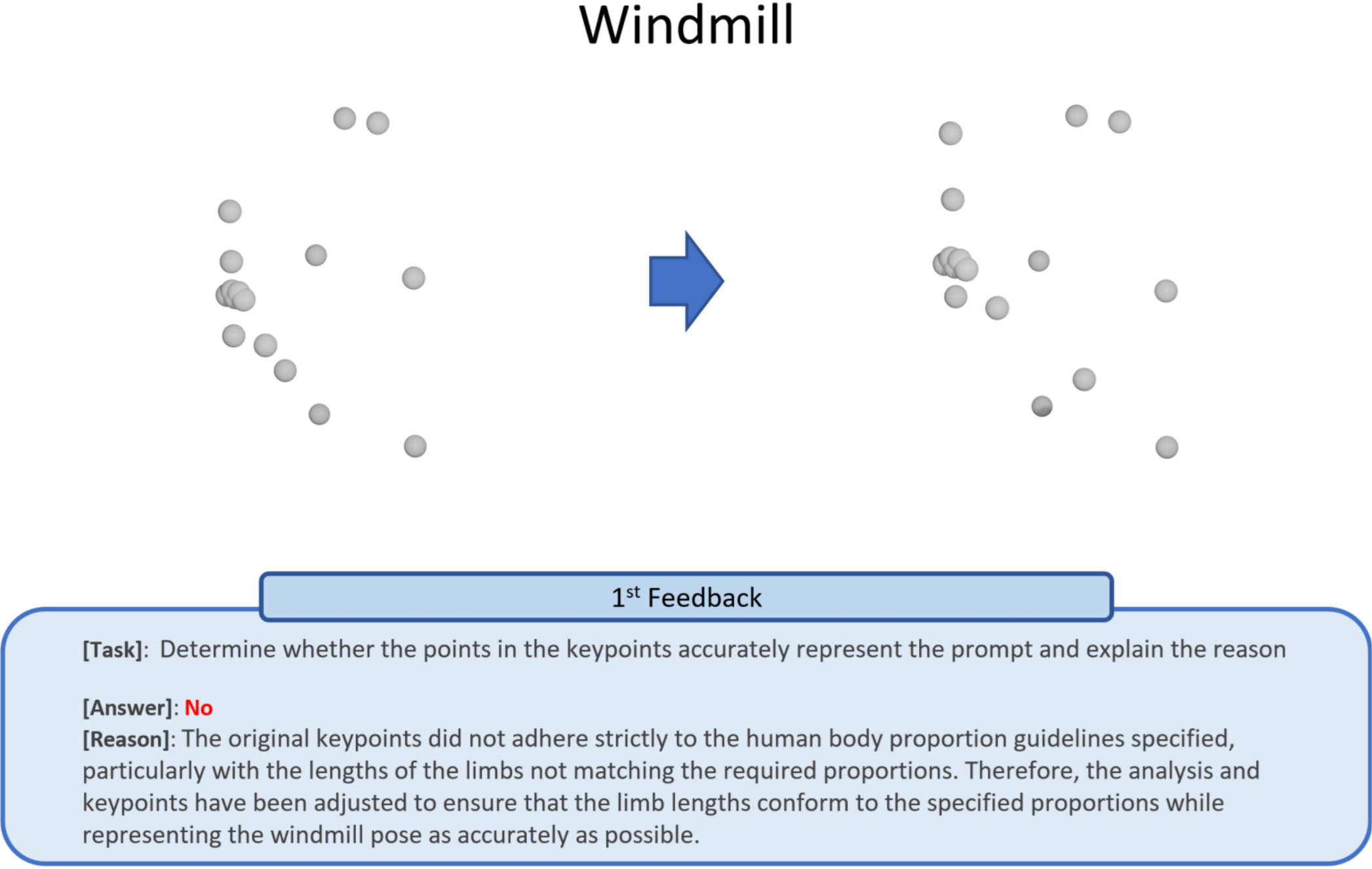}
    \caption{\textbf{First feedback module for the windmill pose.}}
    \label{fig:feedback-sample-acrobatic}
\end{figure}

\begin{figure}[ht]
    \centering
    \vspace{2.5cm}
    \includegraphics[width=1\linewidth]{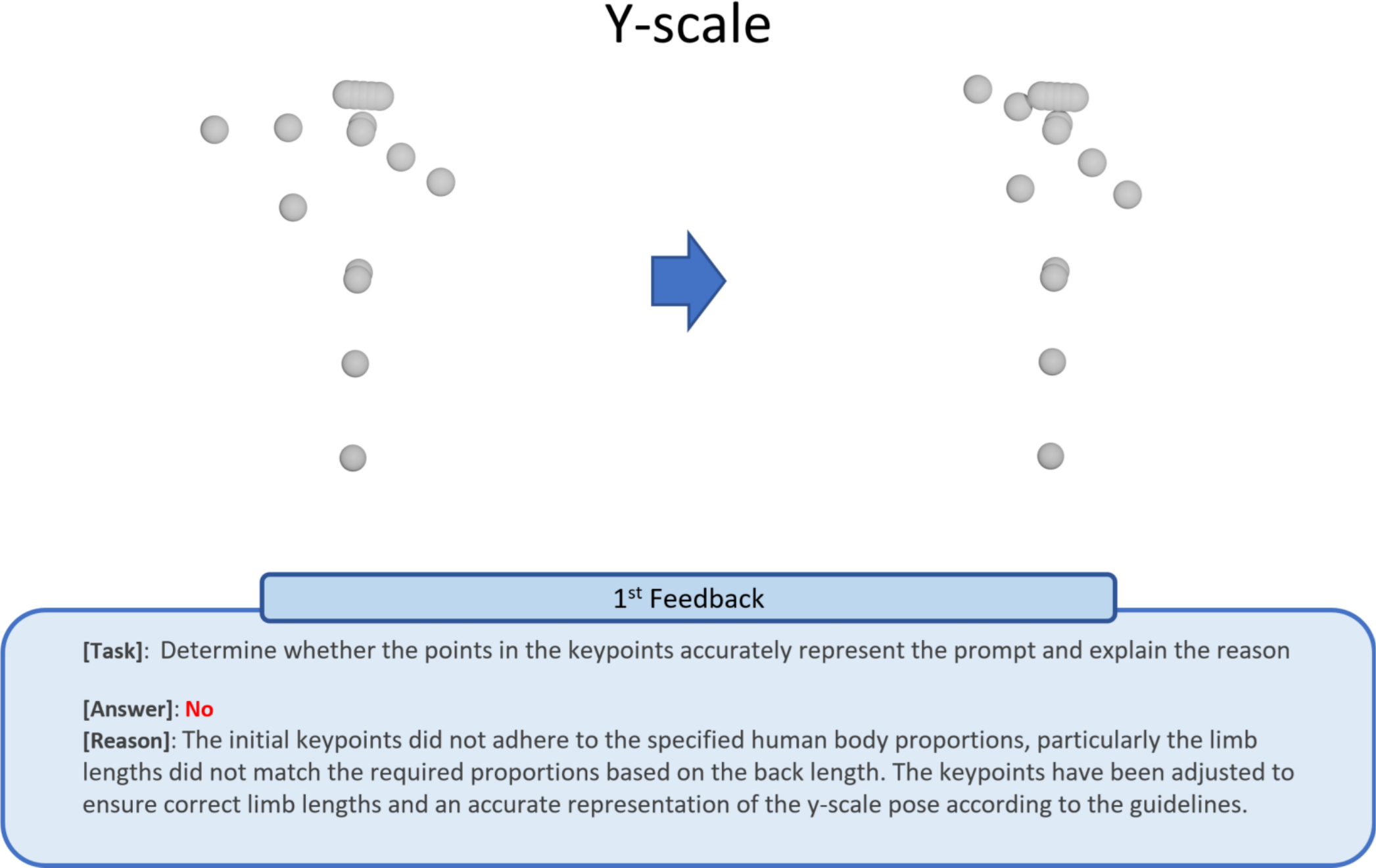}
    \caption{\textbf{First feedback module for the y-scale pose.}}
    \label{fig:feedback-sample-common}
\end{figure}

While the previous results were evaluated based on the final image outputs, we now aim to visualize how the keypoints improve through the feedback process, as illustrated in \Cref{fig:feedback-sample-yoga,fig:feedback-sample-acrobatic,fig:feedback-sample-common}.

\clearpage

\subsection{Robustness to Prompts}
\label{sec:prompt}

SD-based models typically rely on the CLIP model \cite{radford2021learning} to interpret textual prompts. However, CLIP exhibits limitations when processing prompts containing unfamiliar vocabulary \cite{radford2021learning} or lengthy descriptions \cite{zhang2024long} that extend beyond the scope of its training data. Our framework mitigates these limitations by using keypoint information from the prompt as additional conditioning. This enhances robustness to variations in prompt formulation and length. To validate this capability, we conduct experiments using original yoga-pose prompts and prompts rephrased into detailed descriptions of body-part movements. Below, we provide rephrased versions of the yoga pose prompts applied in our experiments.

\begin{itemize}
  \item The Cow and cat pose: The pose by arching their back, lifting their chest and tailbone, and gazing slightly forward while on all fours.
  \item The Downward dog pose: The pose by lifting their hips, straightening their limbs, and forming an inverted V-shape.
  \item The Boat pose: The pose by leaning back slightly, lifting their legs, and extending their arms forward while engaging their core for balance.
  \item The Split pose: The pose sits with one leg split forward and the other split backward along the floor.
  \item The Tree pose: The pose by standing on one leg, placing the other foot against the inner thigh or calf, and bringing their hands together at the chest or reaching overhead while maintaining balance.
\end{itemize}

\begin{figure}[ht] 
    \centering
        \includegraphics[width=1\textwidth]{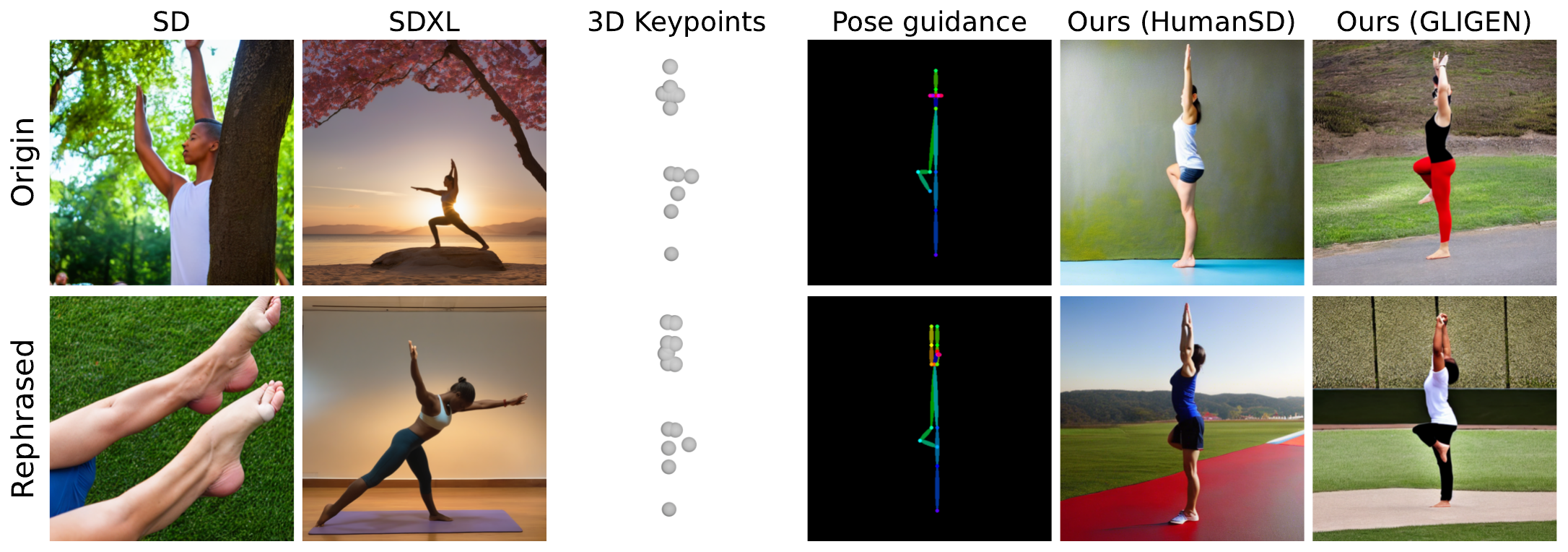}
    \caption{\textbf{Model outputs for original and rephrased prompts, including intermediate stages.} The figure visualizes how original and rephrased prompts affect various stages of the pipeline---3D keypoint, pose guidance, and image generation. Our model effectively captures the same pose described in both the original and rephrased prompts, whereas the baselines often fail to maintain this consistency.}
    \label{fig:reph_}
    \vspace{-8pt}
\end{figure}

\begin{figure}[ht]
    \centering
    \vspace{2.5cm}
    \hspace{-1.7cm}
    \includegraphics[width=1\linewidth]{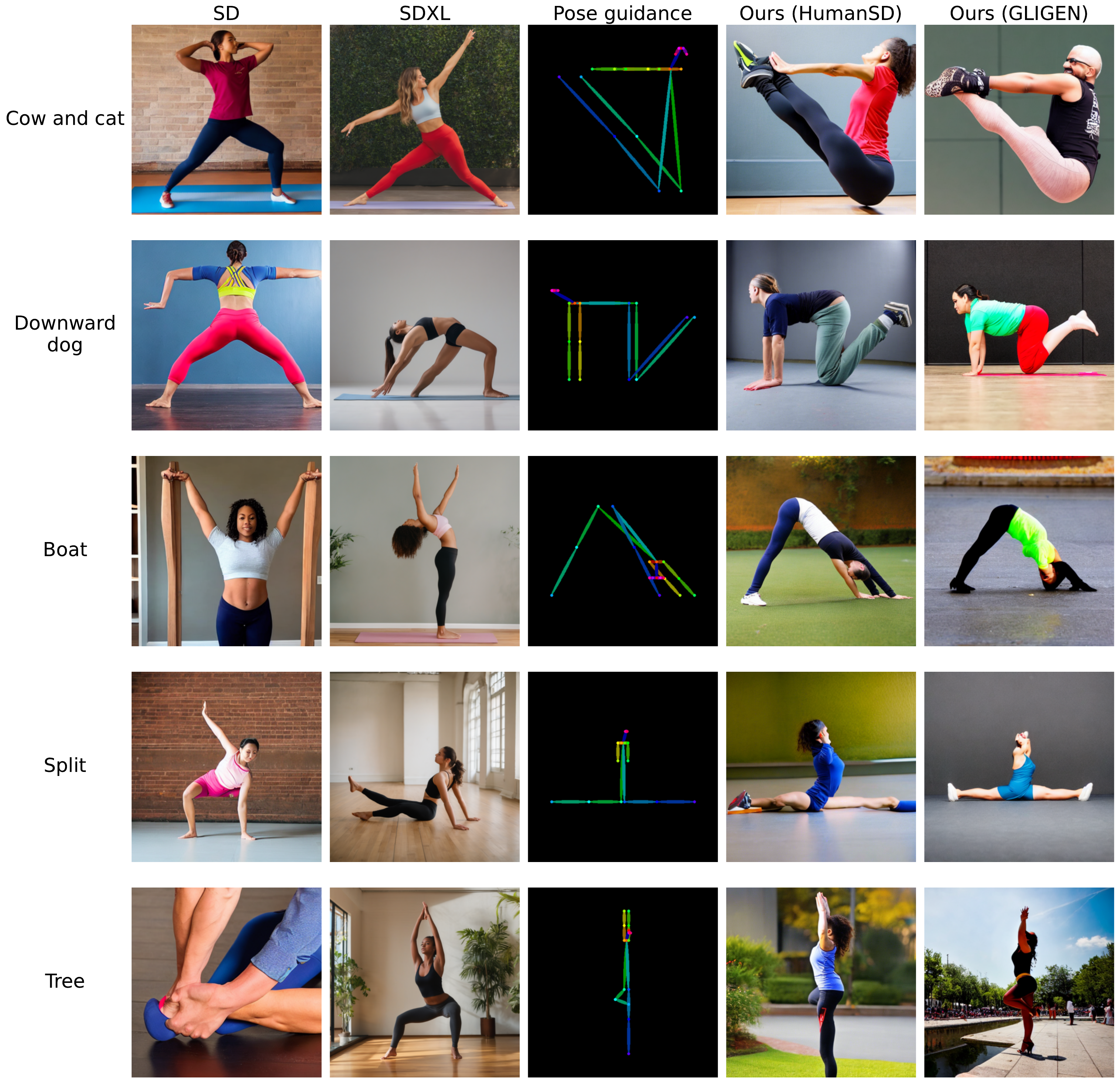}
    \caption{\textbf{Generated images from rephrased yoga prompts.} }
    \label{fig:reph-appen}
\end{figure}

\Cref{fig:reph_} shows the pipeline outputs for the tree pose using both original and rephrased prompts, covering 3D keypoints, pose guidance, and the final image. This observation demonstrates that our framework consistently generates similar keypoint sets from both original and rephrased prompts, confirming its robustness to variations in textual descriptions.
Further examples are provided in \Cref{fig:reph-appen}, which presents images generated for various poses from the rephrased prompt.
Also, as shown in \Cref{tab:abl-reph-scores,tab:abl-reph-acc}, our approach significantly outperforms the baselines across both prompt types, particularly achieving higher VQAScore and accuracy. These findings suggest that our method consistently produces accurate depictions of human poses, regardless of prompt formulation.

\begin{table*}[ht]
    \centering
    
      \scalebox{0.5}{
    \begin{tabular}{ccccccccccccccccccccc}
      \toprule
      Yoga pose & \multicolumn{2}{c}{Cow and cat pose} & \multicolumn{2}{c}{Downward dog pose} & \multicolumn{2}{c}{Boat pose} & \multicolumn{2}{c}{Split pose} & \multicolumn{2}{c}{Tree pose} & \multicolumn{2}{c}{AVG} \\ 
      \cmidrule(lr){2-3}
        \cmidrule(lr){4-5}
        \cmidrule(lr){6-7}
        \cmidrule(lr){8-9}
        \cmidrule(lr){10-11}
        \cmidrule(lr){12-13}
Scores  & \multicolumn{1}{c}{VQA $\uparrow$} & \multicolumn{1}{c}{CLIP $\uparrow$} & \multicolumn{1}{c}{VQA $\uparrow$} & \multicolumn{1}{c}{CLIP $\uparrow$} & \multicolumn{1}{c}{VQA $\uparrow$} & \multicolumn{1}{c}{CLIP $\uparrow$} & \multicolumn{1}{c}{VQA $\uparrow$} 
& \multicolumn{1}{c}{CLIP $\uparrow$} & \multicolumn{1}{c}{VQA $\uparrow$}& \multicolumn{1}{c}{CLIP $\uparrow$} & \multicolumn{1}{c}{VQA $\uparrow$} & \multicolumn{1}{c}{CLIP $\uparrow$}\\
      \midrule
      SD 1.4 & 0.174 & 0.221 & 0.001 & \underline{0.230} & 0.001 & \underline{0.227} & 0.116 & 0.228 & 0.001 & 0.220 & 0.058 & 0.225\\
      SDXL & 0.186 & \textbf{0.239} & 0.057 & 0.229 & 0.001 & \textbf{0.237} & 0.001 & \underline{0.232} & 0.111 & \underline{0.234} & 0.071 & \underline{0.234} \\
      \midrule
      Ours (HumanSD) & \underline{0.652} & \underline{0.235} & \textbf{0.590} & \textbf{0.263} & \underline{0.835} & 0.225 & \underline{0.930} & \textbf{0.253} & \textbf{0.763} & \textbf{0.250} & \underline{0.754} & \textbf{0.245}  \\
      Ours (GLIGEN) & \textbf{0.897} & 0.232 & \underline{0.556} & 0.229 & \textbf{0.933} & 0.217 & \textbf{0.932} & 0.231 & \underline{0.685} & 0.226 & \textbf{0.801} & 0.227   \\      
      \bottomrule
    \end{tabular}
    }
    \caption{ \textbf{VQAScore and CLIPScore for the rephrased prompts.} For brevity, we refer to VQAScore and CLIPScore as VQA and CLIP, respectively. \textbf{Bold} : Best, \underline{underline} : second best. }
    \label{tab:abl-reph-scores}
\end{table*}

\begin{table*}[ht]
    \centering
    
        \scalebox{0.65}{
    \begin{tabular}{ccccccccccccccccccccc}
      \toprule
      Yoga pose & {Cow and cat pose}  & {Downward dog pose} &{Boat pose} & {Split pose} & {Tree pose} & {AVG} \\ 

      \midrule
      SD 1.4 & 81.2\% & 0.00\% & 18.8\% & 56.3\% & 31.2\% & 37.5\% \\ 
      SDXL & 68.7\% & 62.5\% & 25.0\% & 62.1\% & 81.3\% & 59.9\%\\ 
      \midrule
        Ours (HumanSD) & \underline{93.3\%} & \textbf{86.7\%} & \textbf{86.7\%} & \textbf{100\%} & \textbf{100\%} & \underline{93.3\%} \\
      Ours (GLIGEN) & \textbf{100\%} & \underline{85.3\%} & \underline{85.3\%} & \textbf{100\%} & \textbf{100\%} & \textbf{94.1\%}  \\

      \bottomrule
    \end{tabular}
    }
    \caption{\textbf{Classification accuracy for the rephrased prompts.} \textbf{Bold} : Best, \underline{underline} : second best.}
    \label{tab:abl-reph-acc}
    
\end{table*}

\clearpage

\subsection{Compatibility with Various Image Generators}
\label{sec:compatibility}
To demonstrate that our framework can effectively generalize to a variety of image generators, we additionally evaluate its performance using ControlNet \cite{controlnet} and T2I-adapter \cite{t2i}, two widely-used SD-based models capable of utilizing pose conditions. For these experiments, we render skeleton images from the keypoints extracted from prompts, following the procedure used by OpenPose \cite{cao2019openpose, simon2017hand, cao2017realtime,wei2016convolutional}. These skeleton images are subsequently employed as conditioning inputs to ControlNet and T2I-adapter for the final image generation process. 

\begin{figure*}[ht] 
    \centering
        \includegraphics[width=1\textwidth]{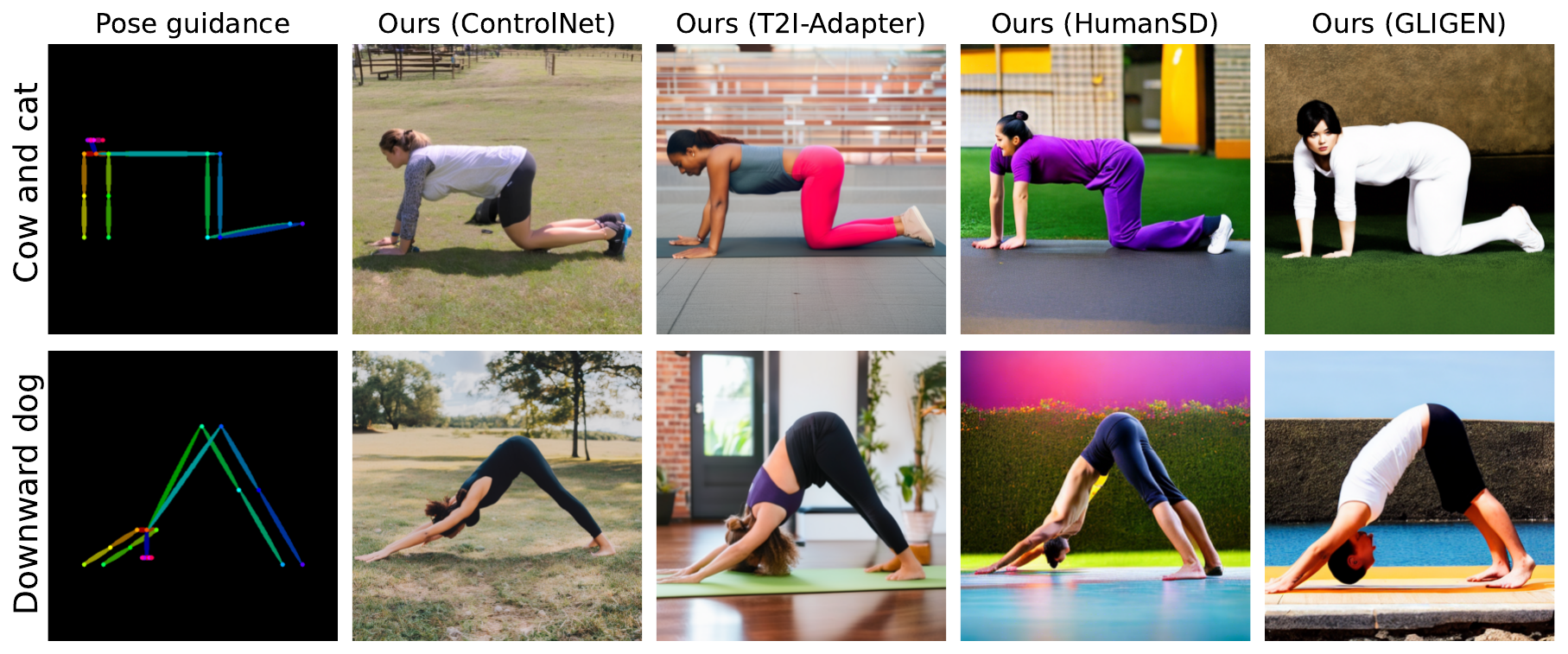}
    \caption{ \textbf{Qualitative results demonstrating the compatibility of our method with various image generators.} The leftmost column shows pose guidance followed by generated samples from four different backbones (ControlNet, T2I-Adapter, HumanSD, and GLIGEN).}
        
    \label{fig:t2i_control_pose_comparison1}
    \vspace{-8pt}
\end{figure*}

\begin{figure}[ht]
    \centering
    \vspace{2.5cm}
    \hspace{-1.7cm}
    \includegraphics[width=1\linewidth]{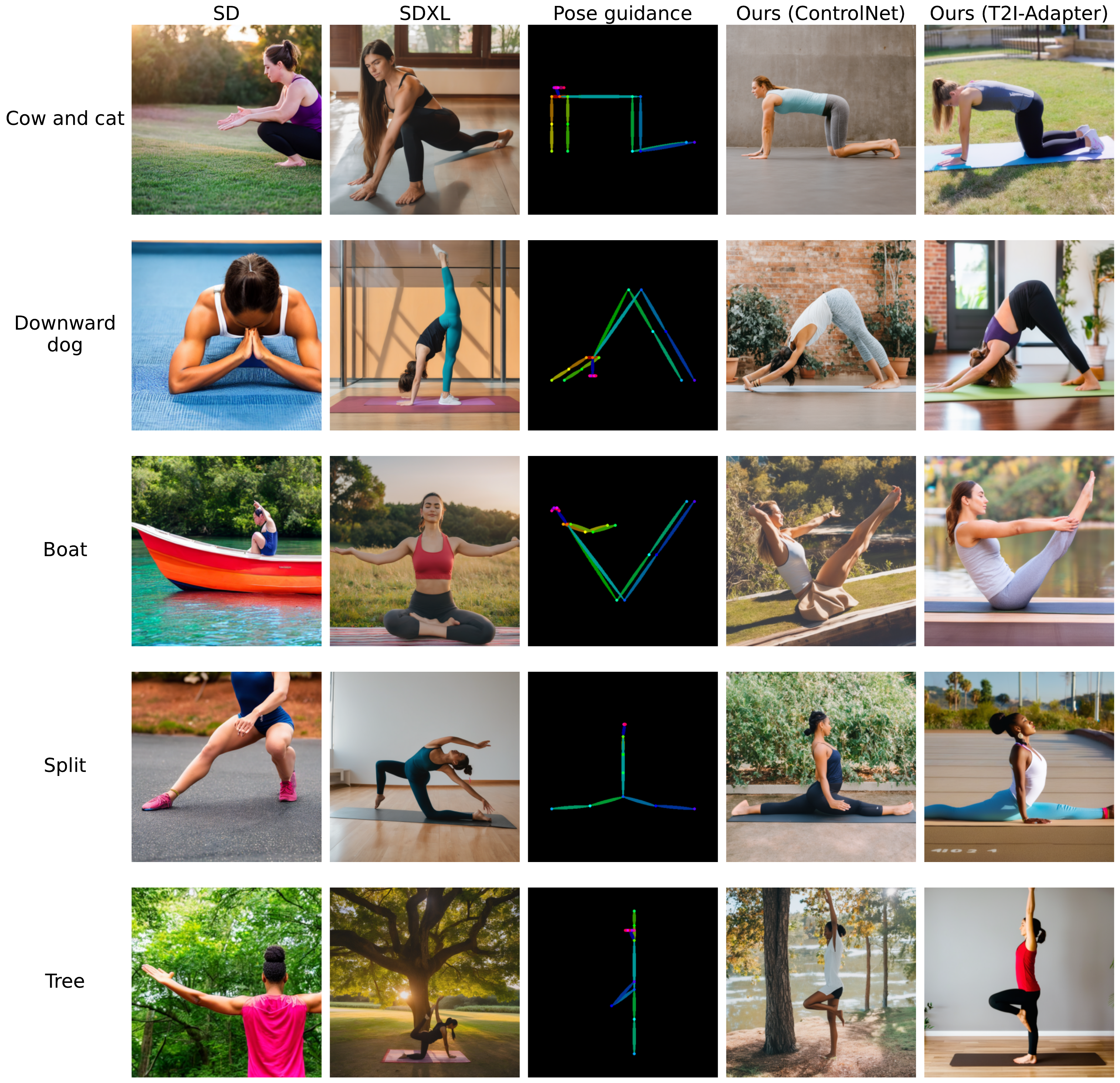}
    \caption{\textbf{Generated images from various generators (ControlNet, T2I-Adapter).}}
    \label{fig:various-gen-appen}
\end{figure}

\Cref{fig:t2i_control_pose_comparison1} presents qualitative examples of yoga poses produced using various image generators, demonstrating visual consistency. Despite the differences in image generators, our method consistently produces pose-accurate images, indicating its flexibility in handling diverse generators. In \Cref{fig:various-gen-appen}, we present additional results for yoga poses.

Quantitative results in \Cref{tab:abl-vqa-clip,tab:abl-accuracy} demonstrate the flexibility of our method across different image generators. In \Cref{tab:abl-vqa-clip}, our method outperforms in VQAScore for all yoga poses, indicating superior alignment with the target pose. While CLIPScore remains relatively similar across models, the strong advantage in VQAScore highlights our method's ability to capture pose-specific features more accurately than the baselines. Moreover, as shown in \Cref{tab:abl-accuracy}, our method achieves strong classification performance across a wide range of poses. Both quantitative and qualitative results confirm that our method is compatible with various image generators, consistently achieving high-quality results across diverse poses and evaluation metrics.

\newpage
\begin{table*}[ht]
    \centering
    
      \scalebox{0.5}{
    \begin{tabular}{ccccccccccccccccccccc}
      \toprule
      Yoga pose & \multicolumn{2}{c}{Cow and cat pose} & \multicolumn{2}{c}{Downward dog pose} & \multicolumn{2}{c}{Boat pose} & \multicolumn{2}{c}{Split pose} & \multicolumn{2}{c}{Tree pose} & \multicolumn{2}{c}{AVG} \\ 
      \cmidrule(lr){2-3}
        \cmidrule(lr){4-5}
        \cmidrule(lr){6-7}
        \cmidrule(lr){8-9}
        \cmidrule(lr){10-11}
        \cmidrule(lr){12-13}
Scores  & \multicolumn{1}{c}{VQA $\uparrow$} & \multicolumn{1}{c}{CLIP $\uparrow$} & \multicolumn{1}{c}{VQA $\uparrow$} & \multicolumn{1}{c}{CLIP $\uparrow$} & \multicolumn{1}{c}{VQA $\uparrow$} & \multicolumn{1}{c}{CLIP $\uparrow$} & \multicolumn{1}{c}{VQA $\uparrow$} 
& \multicolumn{1}{c}{CLIP $\uparrow$} & \multicolumn{1}{c}{VQA $\uparrow$}& \multicolumn{1}{c}{CLIP $\uparrow$} & \multicolumn{1}{c}{VQA $\uparrow$} & \multicolumn{1}{c}{CLIP $\uparrow$}\\
      \midrule
      SD 1.4 & 0.097 & \underline{0.235} & 0.153 & 0.240 & 0.001 & \textbf{0.236} & 0.133 & 0.228 & 0.055 & \underline{0.245} & 0.088 & \underline{0.237}\\
      SDXL & 0.079 & \textbf{0.243} & 0.298 & \underline{0.244} & 0.001 & \underline{0.222} & 0.279 & 0.233 & 0.438 & \textbf{0.256} & 0.219 & \textbf{0.240} \\
      \midrule
      Ours (ControlNet) & \textbf{0.429} & 0.225 & \underline{0.457} & 0.229 & \underline{0.243} & 0.205 & \textbf{0.997} & \textbf{0.248} & \textbf{0.785} & 0.225 & \textbf{0.582} & 0.226 \\
      Ours (T2I-Adapter) & \underline{0.206} & 0.233 & \textbf{0.555} & \textbf{0.251} & \textbf{0.472} & 0.216 & \underline{0.988} & \underline{0.244} & \underline{0.509} & 0.224 & \underline{0.546} & 0.234 \\      
      \bottomrule
    \end{tabular}
    }
    \caption{ \textbf{VQAScore and CLIPScore across various image generators.} For brevity, we refer to VQAScore and CLIPScore as VQA and CLIP, respectively. \textbf{Bold} : Best, \underline{underline} : second best.}
    \label{tab:abl-vqa-clip}

  \vspace{3em}   

    \centering
    
        \scalebox{0.65}{
    \begin{tabular}{ccccccccccccccccccccc}
      \toprule
      Yoga pose & {Cow and cat pose}  & {Downward dog pose} &{Boat pose} & {Split pose} & {Tree pose} & {AVG} \\ 
    \midrule
      SD 1.4 & 31.3\% & 37.5\% & 6.2\% & 50.0\% & 75.0\% & 40.0\%\\
      SDXL & \underline{68.7\%} & \underline{81.8\%} & 18.8\% & 68.8 \% & \underline{87.5\%} & 65.1\% \\
      \midrule
      
      Ours (ControlNet)  & \textbf{73.4\%} & \textbf{86.6\%} & \underline{48.1\%} & \textbf{100.0\%} & \textbf{100.0\%} & \textbf{81.6\%} \\ 
      Ours (T2I-adapter) & 50.6\% & 81.3\% & \textbf{85.2\%} & \underline{98.2\%} & 85.1\% &  \underline{80.8\%} \\

      \bottomrule
    \end{tabular}
    }
    \caption{\textbf{Classification accuracy across various image generators.} \textbf{Bold} : Best, \underline{underline} : second best.}
    \label{tab:abl-accuracy}
    
\end{table*}

\clearpage

\subsection{Compatibility with Various LLMs}
To evaluate the applicability of our framework across different LLMs, we conduct experiments with models such as ChatGPT-o3 \cite{OpenAI2024o3}  and Gemini 2.5 pro \cite{gemini2025_2.5pro}. To assess performance under varying levels of complexity, we select pose samples from yoga, acrobatic, and normal pose categories.

\begin{figure*}[ht] 
    \centering
        \includegraphics[width=1\textwidth]{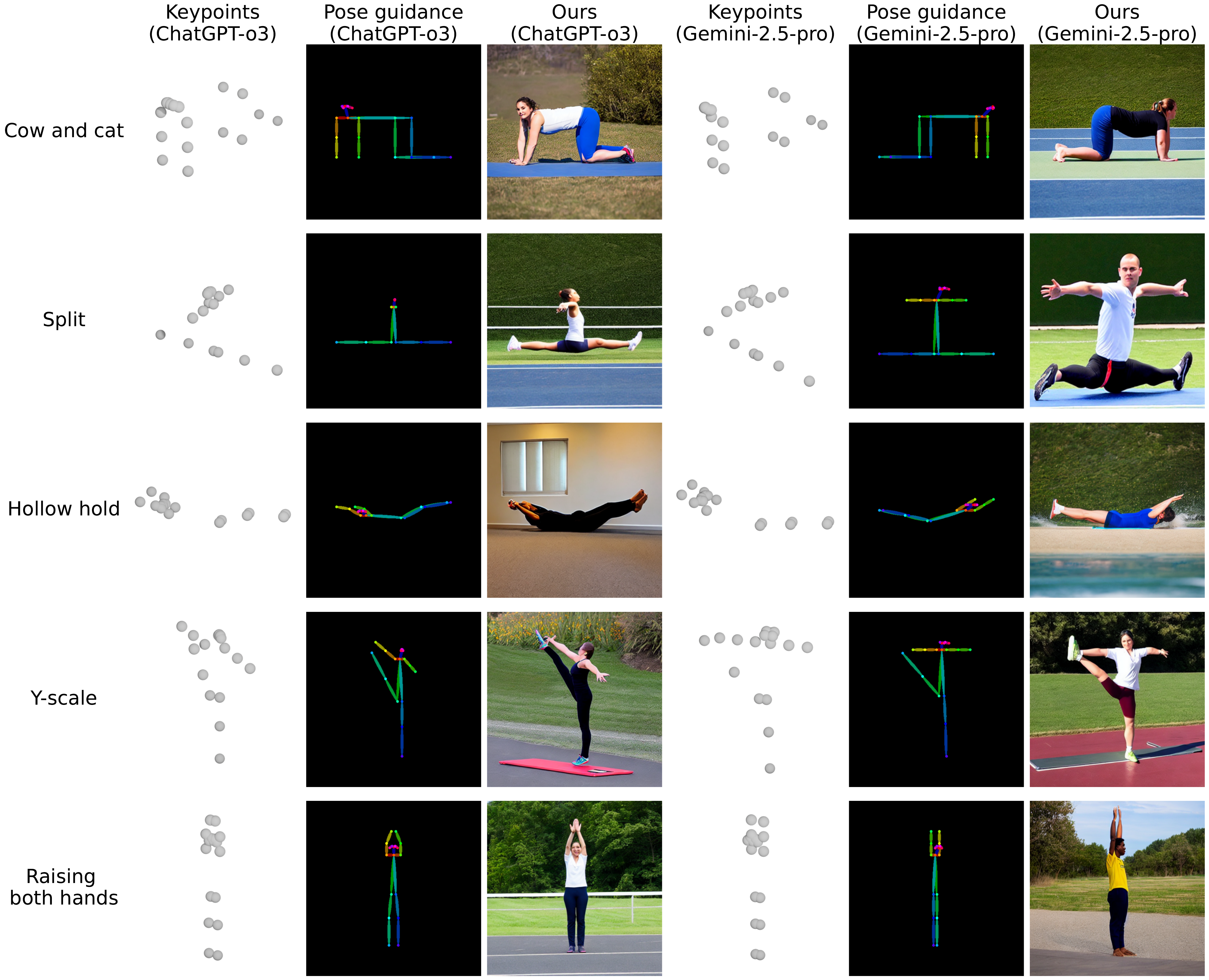}
    \caption{\textbf{Generated images from various LLMs (ChatGPT-o3, Gemini-2.5-pro)}, using GLIGEN as the image generator.}
        
    \label{fig:comparableLLM}
    \vspace{-8pt}
\end{figure*}

As reported in \Cref{tab:abl-LLMs}, our framework achieves consistently strong results across different LLMs, comparable to or even surpassing the original setup. Notably, the variant employing ChatGPT-o3 demonstrates superior performance in several cases, particularly for yoga poses. \Cref{fig:comparableLLM} presents qualitative results across various LLMs, illustrating that our framework consistently produces accurate keypoints and realistic pose-conditioned images.

Overall, these results confirm that our framework is not limited to a particular LLM. Instead, we find that our framework operates reliably and effectively across a diverse set of LLMs, consistently delivering strong performance independent of the specific LLM employed.

\begin{table}[h]
    \centering
      \scalebox{0.5}{
    \begin{tabular}{ccccccccccccccccccccc}
      \toprule
       \multirow{2}{*}{Various pose} & \multicolumn{4}{c}{Yoga pose} & \multicolumn{4}{c}{Acrobatic pose} & \multicolumn{2}{c}{Normal pose} & \multicolumn{2}{c}{\multirow{2}{*}{AVG}} \\
       & \multicolumn{2}{c}{Cow and cat pose} & \multicolumn{2}{c}{Split pose} & \multicolumn{2}{c}{Hollow hold pose} & \multicolumn{2}{c}{Y-scale pose} & \multicolumn{2}{c}{Raising both hands} & & \\ 
      \cmidrule(lr){2-3}
        \cmidrule(lr){4-5}
        \cmidrule(lr){6-7}
        \cmidrule(lr){8-9}
        \cmidrule(lr){10-11}
        \cmidrule(lr){12-13}
Scores  & \multicolumn{1}{c}{VQA $\uparrow$} & \multicolumn{1}{c}{CLIP $\uparrow$} & \multicolumn{1}{c}{VQA $\uparrow$} & \multicolumn{1}{c}{CLIP $\uparrow$} & \multicolumn{1}{c}{VQA $\uparrow$} & \multicolumn{1}{c}{CLIP $\uparrow$} & \multicolumn{1}{c}{VQA $\uparrow$} 
& \multicolumn{1}{c}{CLIP $\uparrow$} & \multicolumn{1}{c}{VQA $\uparrow$}& \multicolumn{1}{c}{CLIP $\uparrow$} & \multicolumn{1}{c}{VQA $\uparrow$} & \multicolumn{1}{c}{CLIP $\uparrow$}\\
      \midrule
      SD 1.4 & 0.097 & 0.235 & 0.133 & 0.228 & 0.001 & 0.191 & 0.001 & 0.197 & 0.778 & {0.221} & 0.202 & 0.214 \\
      SDXL & 0.079 & 0.243 & 0.279 & 0.233 & 0.001 & \underline{0.191} & 0.054 & 0.218 & 0.768 & 0.226 & 0.236 & 0.222 \\
      \midrule

      Ours (o3) & \textbf{0.954} & \textbf{0.257} & \textbf{1.000} & \underline{0.256} & \underline{0.993} & {0.190} & \textbf{0.999} & \underline{0.224} & \underline{1.000} & \underline{0.226} & \textbf{0.989} & \underline{0.231} \\   
      Ours (Gemini 2.5 pro) & {0.859} & 0.245 & {0.999} & \textbf{0.257} & {0.642} & 0.179 & {0.803} & {0.212} & \textbf{1.000} & 0.220 & {0.861} & 0.222 \\   
      \midrule
    Ours (default) & \underline{0.926} & \underline{0.250} & \underline{0.999} & 0.251 & \textbf{0.999} & \textbf{0.211} & \underline{0.989} & \textbf{0.226} & 0.996 & \textbf{0.227} & \underline{0.982} & \textbf{0.233} \\
      \bottomrule
    \end{tabular}
    }
    \caption{ \textbf{VQAScore and CLIPScore across various keypoint generators.} We use GLIGEN as an image generator. For brevity, we refer to VQAScore and CLIPScore as VQA and CLIP, respectively. \textbf{Bold} : Best, \underline{underline} : second best.}
    \label{tab:abl-LLMs}
    
\end{table}

\clearpage

\subsection{Sensitivity Analysis for Prompt Template }
In this section, we examine the sensitivity of the prompt template serving as conditional input to the LLM. To assess this effectively, we investigate how the absence of each component in the template affects the final output of our framework.

We focus on the following three elements: (1) Grounding condition (Guidelines 1-3), which defines the body’s spatial relation to the ground by specifying where contact points are positioned; (2) Human proportion (Guidelines 4), which imposes anatomical constraints on relative body part sizes; and (3) Process steps, which instruct the model to extract pose-relevant information from the prompt. As we describe in \Cref{subsec:imple_detail}, the prompt template consists of two main parts: keypoint generation specification and base instruction. All three of the above elements belong to the keypoint generation specification. Other elements are excluded from this analysis, as their removal either makes the template internally inconsistent or prevents the LLM from producing structured outputs. The template structure is shown below.

{\scriptsize
\[
\begin{array}{l}
\begin{array}{l}
\text{Keypoint generation} \\
\text{\qquad specification}
\end{array}
\left\{ \begin{array}{l}
\text{Objective} \\
\text{...} \\
\text{Keypoint Format}\\
\text{...} \\
\text{Guidelines}
\left\{
\begin{array}{l}
\textbf{1-3: Grounding condition}\\
\textbf{4: Human proportion}
\end{array}
\right.\\
\text{...} \\
\textbf{Process steps}\\
\end{array} \right. \\[5mm]
\text{\qquad Base instruction\quad} \left\{ \begin{array}{l}
\\
\text{Instruction for task} \\
\\
\end{array} \right.
\end{array}
\]
}

\begin{table}[ht]
    \centering
    \scalebox{0.55}{
    \begin{tabular}{ccccccccccccccccccccc}
      \toprule
      Acrobatic pose & \multicolumn{2}{c}{Hollow hold} & \multicolumn{2}{c}{L-sit} & \multicolumn{2}{c}{One-arm handstand} & \multicolumn{2}{c}{Windmill} & \multicolumn{2}{c}{Y-scale} & \multicolumn{2}{c}{AVG} \\ 
      \cmidrule(lr){2-3}
      \cmidrule(lr){4-5}
      \cmidrule(lr){6-7}
      \cmidrule(lr){8-9}
      \cmidrule(lr){10-11}
      \cmidrule(lr){12-13}
    
  Template w/o  & \multicolumn{1}{c}{VQA\ $\uparrow$} & \multicolumn{1}{c}{CLIP\ $\uparrow$} & \multicolumn{1}{c}{VQA\ $\uparrow$} & \multicolumn{1}{c}{CLIP\ $\uparrow$} & \multicolumn{1}{c}{VQA\ $\uparrow$}
& \multicolumn{1}{c}{CLIP\ $\uparrow$} & \multicolumn{1}{c}{VQA\ $\uparrow$} & \multicolumn{1}{c}{CLIP\ $\uparrow$} & \multicolumn{1}{c}{VQA\ $\uparrow$} & \multicolumn{1}{c}{CLIP\ $\uparrow$} & \multicolumn{1}{c}{VQA\ $\uparrow$} & \multicolumn{1}{c}{CLIP\ $\uparrow$}\\
      \midrule
       Grounding cond. & 0.225 & 0.181 & \underline{0.622} & 0.207 & 0.294 & 0.252 & 0.229 & 0.225 & 0.001 & 0.218 & 0.274 & 0.217 \\
       Process steps& 0.593 & 0.183 & 0.389 & 0.195 & 0.662 & 0.256 & 0.333 & \underline{0.236} & 0.006 & \underline{0.225} & 0.397 & 0.219 \\
      Human proportion & \underline{0.742} & \underline{0.191} & 0.417 & \underline{0.214} & \underline{0.700} & \underline{0.258} & \underline{0.883} & 0.229 & \underline{0.682} & 0.215 & \underline{0.685} & \underline{0.222} \\      
      
      \midrule
      Full template & \textbf{0.999} & \textbf{0.211} & \textbf{0.807} & \textbf{0.216} & \textbf{0.831} & \textbf{0.261} & \textbf{0.960}  & \textbf{0.241} & \textbf{0.989} & \textbf{0.226} & \textbf{0.918} & \textbf{0.231} \\
      \bottomrule
    \end{tabular}
    }
    \caption{ \textbf{Sensitivity analysis for prompt template.} \textbf{Bold} : Best, \underline{underline} : second best.}
    \label{tab:abl_prompt_template}
\end{table}

\begin{figure*}[t] 
    \centering
        \includegraphics[width=0.6\textwidth]{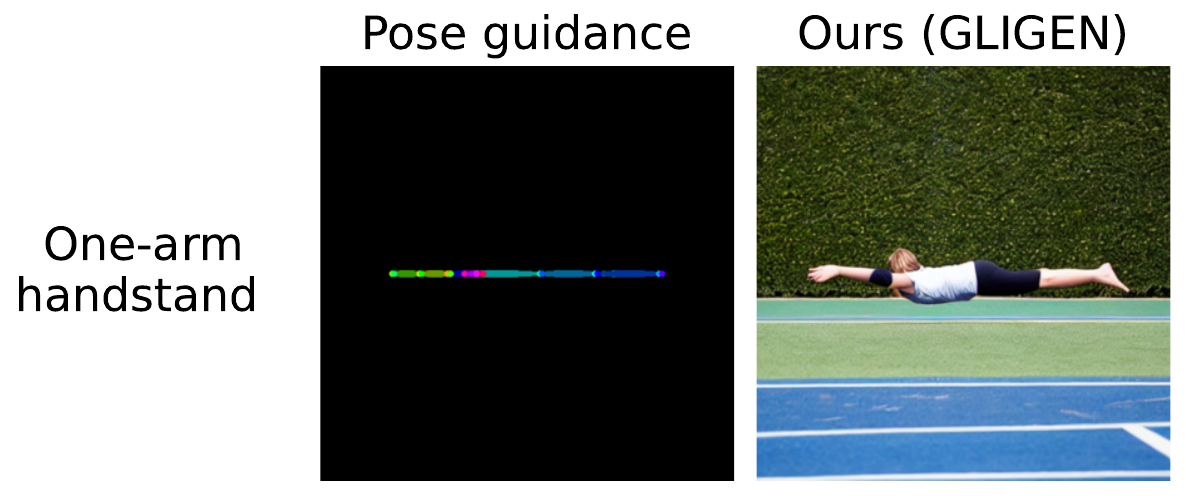}
    \caption{ \textbf{Impact of the grounding condition on the template.}}
        
    \label{fig:able_sensitivity_template}
    \vspace{-8pt}
\end{figure*}

\Cref{tab:abl_prompt_template} presents the results of our prompt sensitivity analysis. The full template yields the best performance overall. Interestingly, removing the human proportion constraint results in better performance than removing the grounding condition or process steps. This suggests that, although the human proportion constraint does influence the output, its impact is relatively limited in comparison. On the other hand, removing the grounding condition leads to the poorest performance across various poses. As illustrated in \Cref{fig:able_sensitivity_template}, this is particularly evident in cases such as the one-arm handstand, where the absence of a ground reference causes the generated pose to appear more as lying down rather than standing upright.

These findings confirm that our prompt template, as a form of conditional input, is well-designed for generating plausible human poses. Furthermore, the results allow us to assess the contribution of each component to the overall pose generation quality, highlighting their relative importance within the template structure. These insights also offer practical guidance for prompt engineering in pose-conditioned generation tasks.

\clearpage 

\section{Discussion} 
\begin{figure*}[ht] 
    \centering
        \includegraphics[width=1\textwidth]{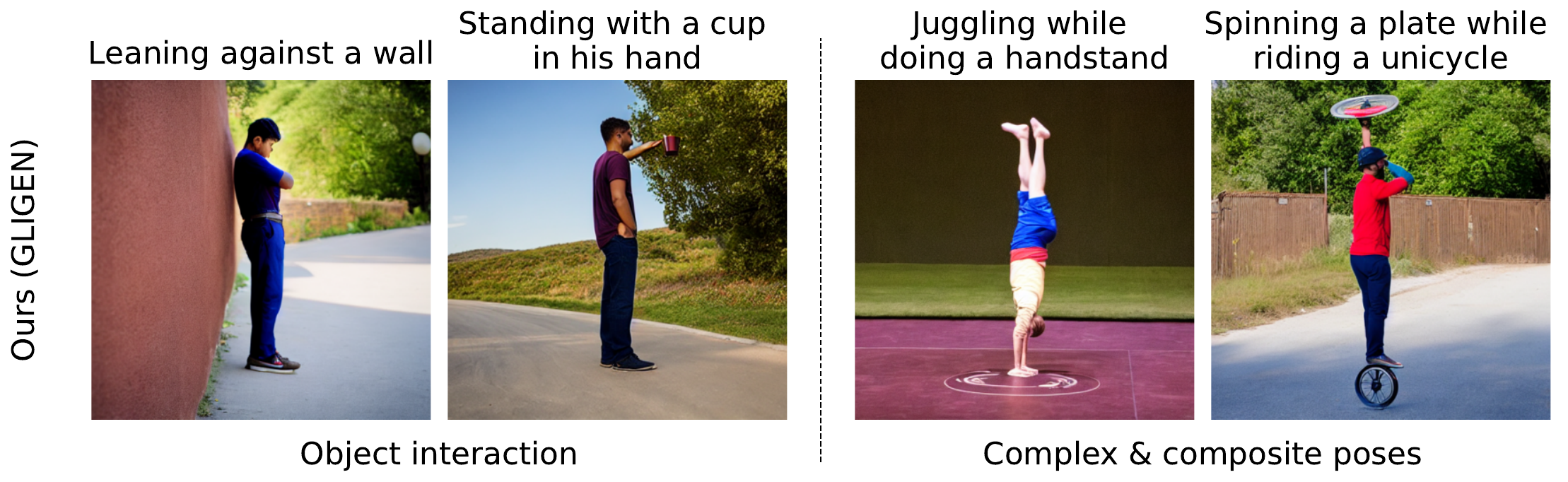}
    \caption{ \textbf{Generated images for challenging prompts.}}
        
    \label{fig:abl-extreme}
    \vspace{-8pt}
\end{figure*}

\paragraph{More Challenging Prompt} To investigate the limitations of our framework, we conducted further experiments in more challenging scenarios. \Cref{fig:abl-extreme} illustrates the results for two pose categories: object interaction and complex composite poses. In the case of object interaction, the model generates images where the human correctly interacts with objects such as the \textit{"wall"} or \textit{"cup"}. On the other hand, for complex composite poses, such as \textit{"juggling while performing a handstand"}, the model fails to produce correct actions. These examples highlight the limitations of our framework, with future research aiming to improve pose guidance for such challenging scenarios.

\begin{figure*}[ht] 
    \centering
        \includegraphics[width=0.65\textwidth]{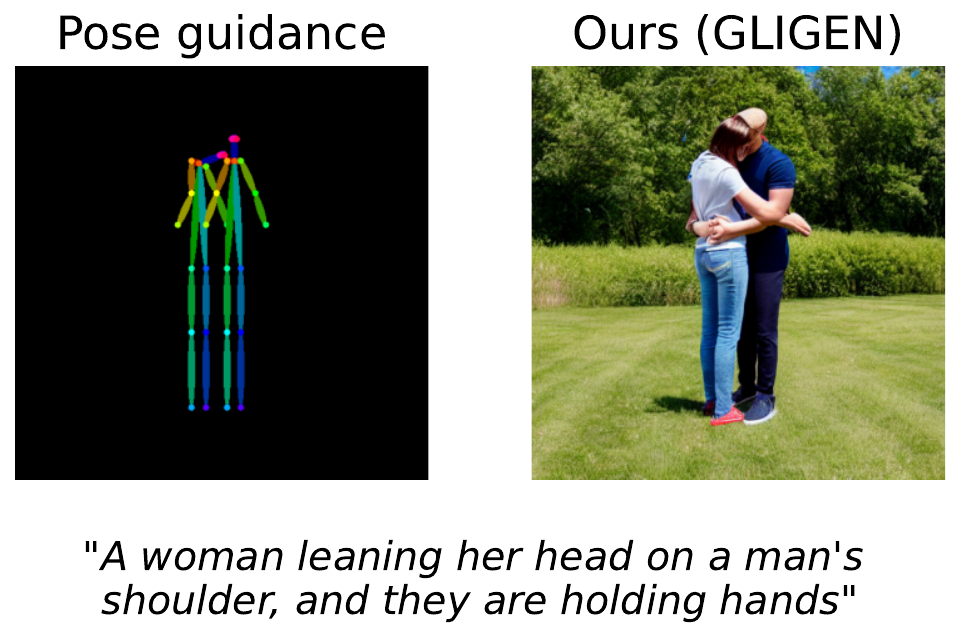}
    \caption{ \textbf{Results for multi-person scenario.}}
        
    \label{fig:abl-multiperson}
    \vspace{-8pt}
\end{figure*}

\paragraph{Multi-person Scenario} Our framework primarily handles cases with a single human. To assess its performance with multiple subjects, we conduct additional experiments with multi-person scenarios. \Cref{fig:abl-multiperson} shows the outputs generated by our framework for multi-person prompts.

The generated samples demonstrate that our framework can produce reasonable outputs in multi-person scenarios, while also revealing both its strengths and limitations. The predicted pose guidance shows a plausible role assignment between the man and woman. Specifically, the coordinates of the keypoints generated are as follows:

\begin{center}
    \textit{Woman 1 - Nose: (-0.05, 0.0, 2.35), Left Eye: (-0.06, 0.0, 2.36), ...} 
    
    \textit{Man 1 - Nose: (-0.15, 0.0, 2.5), Left Eye: (-0.16, 0.0, 2.51), ...}

\end{center} 
The coordinates indicate that the nose and eye positions of the female subject are consistently lower than those of the male subject. The spatial differentiation, where the woman is leaning her head on the man's shoulder, demonstrates that our framework consistently links each pose to the proper individual. From the perspective of spatial and semantic relationships, "leaning" is clearly captured. However, the "holding hands" gesture is not reflected in the pose structure, indicating that multi-person interactions are not always fully represented.

These observations highlight both the potential and current limitations of our framework in handling multi-person scenarios. While our method effectively captures individual poses, it occasionally struggles to handle complex multi-person interactions. Consequently, enhancing the capacity to capture relationships between people remains a key challenge.

\paragraph{Projection Strategy} 
In our framework, the projection with the highest variance is selected as the representative pose guidance image. However, this approach does not always guarantee a semantically meaningful depiction of the action. For instance, in actions such as crouching, where the body folds inward, a high-variance projection may fail to capture the most representative pose. Moreover, the current projection mechanism makes it difficult to specify a projection from a particular viewpoint. To achieve viewpoint-specific projections, it is necessary to ensure that the generated keypoints are consistently aligned in orientation, which poses a significant challenge.

\begin{figure*}[t] 
    \centering
        \includegraphics[width=1\textwidth]{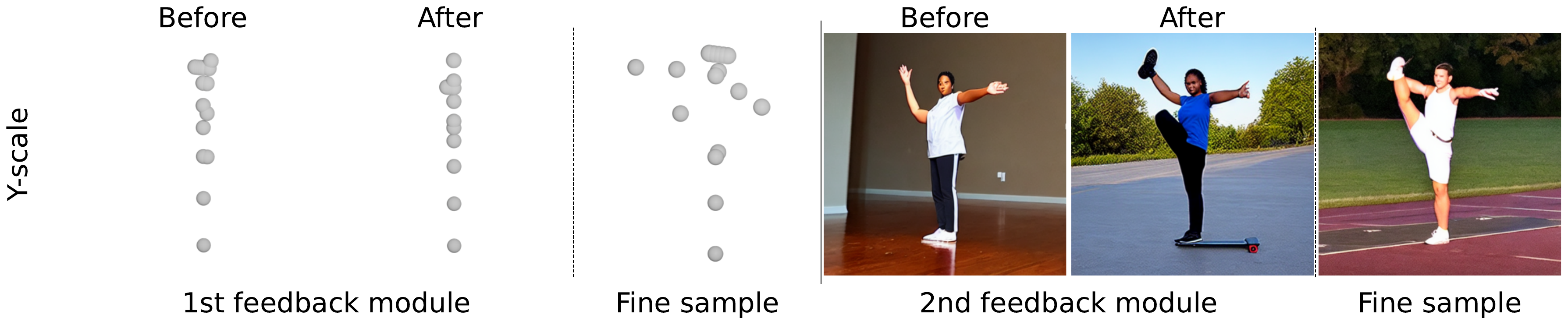}
    \caption{ \textbf{Oscillation of feedback system.}}
        
    \label{fig:able_feedback_fail}
    \vspace{-8pt}
\end{figure*}

\paragraph{Oscillation of Feedback System} The oscillatory behavior of our feedback system---where corrections are attempted but do not consistently lead to improvement---is sometimes observed in practice.
\Cref{fig:able_feedback_fail} shows two representative instances in which the feedback process fails to converge to a satisfactory result.
In the first feedback module, we observe that although the first feedback module correctly detects the inaccuracy of the initial keypoints and attempts to regenerate them, it still fails to produce reasonable keypoints.
Similarly, in the second feedback module, even when the second feedback module identifies the inadequacy of the generated image and generates a new one, the output does not always show clear improvement.
These examples reflect a current limitation of our framework and highlight a direction for future enhancement.

\paragraph{Extension to Non-human Pose} This paper primarily focuses on human poses. To explore how our framework performs on non-human poses, we conduct additional experiments. The default setting refers to a modified version of the original prompt template in which the word “human” is replaced with “object.” In the plain setting, the prompt template retains only the keypoints format---namely, labeled body parts and their 3D coordinates---while removing any textual reference adapted for humans.

\begin{figure*}[ht] 
    \centering
        \includegraphics[width=1\textwidth]{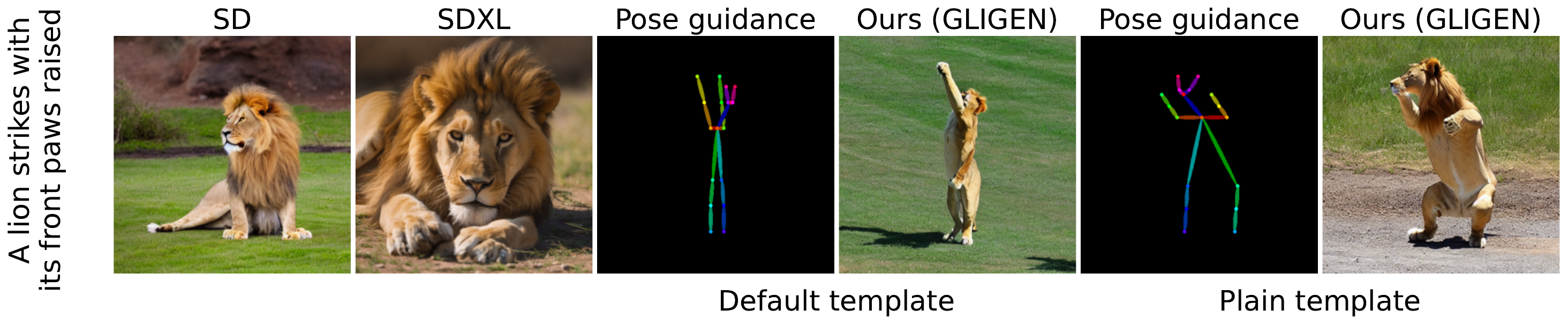}
    \caption{ \textbf{Extension to non-human pose.}}
        
    \label{fig:able_non_human}
    \vspace{-8pt}
\end{figure*}

\Cref{fig:able_non_human} presents the results generated from text prompts related to lions under both the default and plain settings. In both cases, the framework appears to capture the motion to some extent. However, the generated outputs still tend to depict upright, bipedal figures, resembling human poses. This indicates that our current framework struggles to generate appropriate keypoints for non-human entities, and that the image generation module---while faithfully following the given keypoints---ultimately produces outputs biased toward human-like forms.

These observations suggest a fundamental limitation in generalizing from human to non-human domains. We expect that bridging the gap between the prompt, the keypoints (pose guidance), and the final image---particularly for non-human poses---could lead to more semantically accurate and natural image generation.

\paragraph{Computational Cost and Efficiency}
We analyze the computational cost and efficiency of our proposed framework. The measurement is conducted component-wise, following the operational sequence of the framework. \Cref{computational_cost} presents the results. Note that "Pass" or "Fail" in the feedback columns indicates whether the generated outputs align with the pose described in the input prompt. All experiments were conducted on an Nvidia GeForce RTX 3090 GPU. Compared to baseline models, the component-based design of our framework introduces higher latency and efficiency loss, particularly in the keypoint generation stage. These efficiency-related issues remain an important area for further improvement.

\begin{table}[h]
    \centering
    \scalebox{0.6}{
        \begin{tabular}{c|ccc|cccc} 
            \toprule
            \multirow{4}{*}{PointT2I Components} & \multicolumn{3}{c|}{Keypoint generation} & \multicolumn{4}{c}{Image generation}  \\ 
            \cmidrule(lr){2-8}
            & \multirow{2}{*}{Keypoint gen.} & \multicolumn{2}{c|}{1st feedback} & \multirow{2}{*}{Projection} & \multirow{2}{*}{Image gen.} & \multicolumn{2}{c}{2nd feedback} \\ 
            \cmidrule(lr){3-4}
            \cmidrule(lr){7-8}
            & & \multicolumn{1}{c}{Pass} & \multicolumn{1}{c|}{Fail} & & & Pass & Fail \\ 
            \midrule
            SD 1.4 & \multirow{2}{*}{-} & \multirow{2}{*}{-} & \multirow{2}{*}{-} & \multirow{2}{*}{-} & 2.429 & \multirow{2}{*}{-} & \multirow{2}{*}{-} \\ 
            SDXL &  &  &  &  & 13.142 &  &  \\ 
            \midrule
            Ours (HumanSD) & \multirow{2}{*}{76.773} & \multirow{2}{*}{42.118} & \multicolumn{1}{c|}{\multirow{2}{*}{71.840}} & \multirow{2}{*}{0.001} & 6.362 & \multirow{2}{*}{3.117} & 9.482 \\ 
            Ours (GLIGEN) & & & \multicolumn{1}{c|}{} & & 34.042 & & 37.171 \\ 
            \bottomrule
        \end{tabular}
    }
    \caption{\textbf{Computational Cost and Efficiency} (in seconds).}
    \label{computational_cost}
\end{table}


\paragraph{\blue{Compatibility to More Recent Text-to-image Backbones}} 

\begin{figure*}[ht] 
    \centering
        \includegraphics[width=1\textwidth]{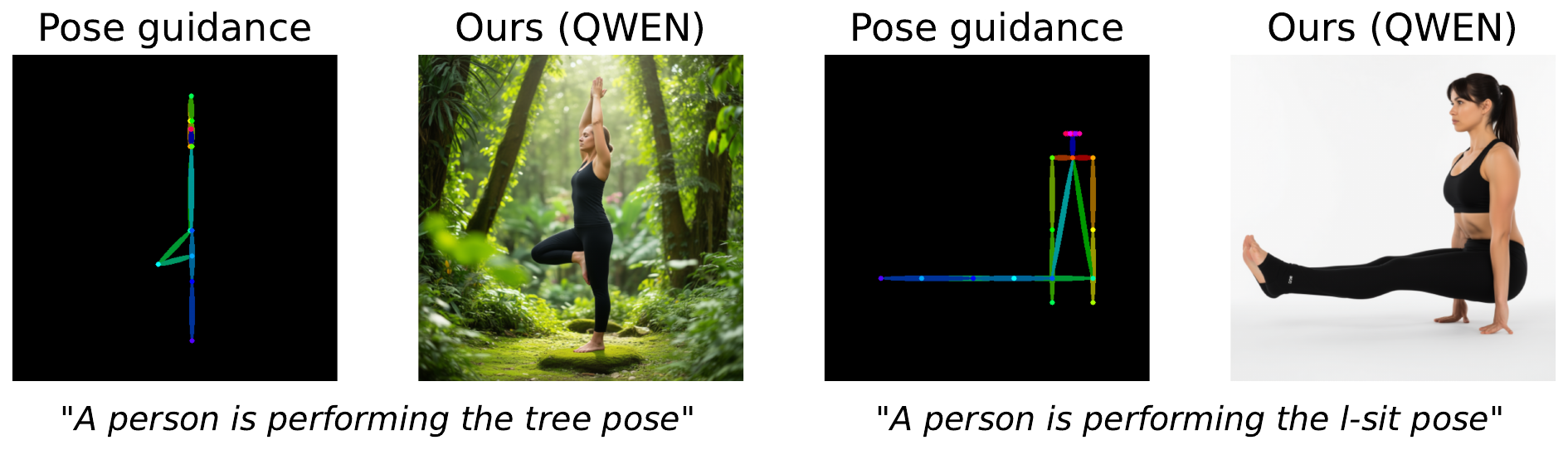}
    \caption{ \textbf{Generated images using a recent image generator (QWEN).}}
        
    \label{fig:qwen}
    \vspace{-8pt}
\end{figure*}

\blue{To further assess the compatibility of our framework with recent T2I models, we conducted additional experiments using QWEN \cite{wu2025qwenimagetechnicalreport}. As shown in \Cref{fig:qwen}, our framework accurately preserves pose structures while maintaining high visual quality---particularly in rendering fine details such as fingers---across various prompts. These results demonstrate the compatibility and robustness of our framework across different T2I generation models. Building on its independence from specific T2I backbones, our framework can be readily extended to more recent models such as FLUX \cite{labs2025flux1kontextflowmatching,flux2024}, SD 3.5 \cite{sd,esser2024scaling}, and QWEN \cite{wu2025qwenimagetechnicalreport}, provided that these models are pretrained with keypoints or skeleton-based images as conditioning inputs.}

\paragraph{Future work}
Building on the limitations discussed above, future work will focus on developing a framework that can more effectively interpret and reflect the input prompts. This includes addressing challenges related to complex actions, prompts involving multi-person scenarios, and modeling human–object interactions. \blue{Another key direction is to enhance the integration of control conditions within diffusion-based architectures, such as pose-aware T2I backbones, moving beyond the current condition-injection strategies to achieve more optimal controllability.} 

\blue{In addition, improving the computational efficiency is an important goal, as the presence of multiple components leads to increased overhead. In particular, the iterative feedback process introduces a clear trade-off between generation quality and speed, which future research should aim to balance more effectively.} 

Furthermore, another important direction is enabling more precise and controllable projection mechanisms, particularly in aligning the generated images with the intended viewpoint. Additional efforts are also needed to overcome the current limitations of the feedback system, such as oscillation, and to extend our framework's ability to handle non-human poses. Advancing these areas will require not only improved prompt engineering strategies but also a deeper understanding of generative models, as well as the ability to effectively leverage large-scale models such as LLMs.

\section{Conclusion}
\label{sec:conclusion}

This work addresses the challenge of generating human poses in text-to-image (T2I) models---an area that demands both high-level semantic understanding and precise structural guidance. We present PointT2I, a framework that bridges text and structure by predicting pose keypoints directly from prompts. In doing so, it enables pose-aware T2I backbones---which require structural inputs but cannot extract them from language alone---to generate pose-accurate images from text.
PointT2I enables accurate pose-conditioned image generation across a wide range of human actions, including uncommon and diverse poses without any pose-specific tuning or external supervision. Its LLM-based feedback system further enhances consistency by refining both the predicted keypoints and the final image. Furthermore, PointT2I handles diverse prompt styles---ranging from explicit pose names to descriptive expressions---while consistently generating structurally aligned images.

\clearpage

\appendix

\section{Instructions to the LLM}
We employ three LLMs in our framework. The first LLM acts as a keypoint generator, producing keypoints that are semantically aligned with the input prompt. The second LLM serves as the first feedback system, evaluating whether the generated keypoints accurately reflect the intended semantics of the prompt. The third LLM functions as the second feedback system, assessing whether the final generated image is consistent with the input prompt.
 
 When designing the LLM template, we established clear guidelines to ensure that each LLM identifies the pose described in the prompt and generates outputs in a structured manner. For example, in the case of keypoint generation by the first LLM, we designated the ground plane to define orientation and generated keypoints sequentially outward from the torso. Each keypoint was placed based on pose-specific constraints and its relation to previously generated points. This sequential generation strategy produced more coherent and semantically accurate keypoints than non-sequential methods.

Now, we present the prompt templates in the following order: (1) Keypoints generator template, (2) 1st feedback system template, and (3) 2nd feedback system template.

\lstset{
    frame=single,                       
    breaklines=true,  
    basicstyle=\footnotesize,
    showstringspaces=False         
}

\subsection{Keypoints generator template}
\label{code:keypoints}

\begin{lstlisting}
# Your Role: Excellent Keypoints Adjuster
## Objective: Manipulate keypoints in square images according to the user prompt while maintaining visual accuracy
## Keypoints specification and Manipulations 

1. Keypoints format: [[x0, y0, z0], #Nose
[x1, y1, z1], # Left Eye
[x2, y2, z2], # Right Eye
[x3, y3, z3], # Left Ear
[x4, y4, z4], # Right Ear
[x5, y5, z5], # Left Shoulder
[x6, y6, z6], # Right Shoulder
[x7, y7, z7], # Left Elbow
[x8, y8, z8], # Right Elbow
[x9, y9, z9], # Left Wrist
[x10, y10, z10], # Right Wrist
[x11, y11, z11], # Left Hip
[x12, y12, z12], # Right Hip
[x13, y13, z13], # Left Knee
[x14, y14, z14], # Right Knee
[x15, y15, z15], # Left Ankle
[x16, y16, z16], # Right Ankle
]
Operation : human body action

## Key Guidelines
0. Analyze: Confirm the position of each keypoint individually to describe the movement among the prompt and Analyze the movement. Consider the hips and sit bones as the same.
1. The "special points": Among the keypoints, find the points that are in contact with the ground. We denote that points as "special points". 
2. Alignment: Follow the user's prompt, keeping the specified movement 
3. The ground is xy plane and a person is located in a space where z >0. The z coordinates of the "special points" must be zero and z coordinates of the other points must be greater than or equal to 0.
4. Human body proportion: The following rules must be adhered to without exception. The upper leg length is 0.6 times the back length. The lower leg length is 0.7 times the back length. The upper arm length is 0.3 times the back length. The lower arm length is 0.3 times the back length. The distance from the midpoint of the shoulders to the nose is 0.2 times the back length.

## Process Steps
1. Interpret prompts: Read and understand the user's prompt. Extract the key features for each action accurately. Make the of the hips identical to the position of the sit bones.
2. Find all the "special points". If there are no "special points", set the point closest to the ground as the "special point". The z-coordinate of the "special points" must be exactly 0.
3. Analyze the pose of each object and organize the information about the found features, stating "This keypoint should be in this position." Format your analysis results as shown in the examples.
4. Generate the position of the hips first. The distance between the left hip and the right hip should be small.
5. Determine the position of the shoulders based on whether the torso is standing or lying down while adopting the pose. Determine whether the torso is lying down or not. Make sure to confirm the position of the torso clearly. If the pose is not a lying down posture, the change in the z-axis between the shoulders and hips should be significant, while the x and y coordinates of the two points should be set to the same value.
6. Considering the fact that a person taking a specific pose is looking to the left, generate the position of the head.
7. Considering the position, tilt, and angle of the legs of a person taking a specific pose, generate the position of the knees so that they are well connected to the hips. It must be generated so that it is clear whether the x-axis, y-axis and z-axis coordinates are smaller or larger than those of the shoulders and hips, respectively. Make sure whether both legs are separated from each other or not. Check the distance between the left knee and the left hip. It should be same as the distance between the right knee and the right hip.
8. Considering the position, tilt, and angle of the legs of a person taking a specific pose, generate the position of the ankles so that they are well connected to the knees. It must be generated so that it is clear whether the x-axis, y-axis and z-axis coordinates are smaller or larger than those of the shoulders, hips, and knees, respectively. The distance between the left knee and the left ankle should be similar to the distance between the left hip and the left knee. The same should apply to the right side as well.
9. Considering the position, tilt, and angle of the arms of a person taking a specific pose, generate the position of the elbows so that they are well connected to the shoulders. It must be generated so that it is clear whether the x-axis, y-axis and z-axis coordinates are smaller or larger than those of the shoulders, hips, knees, and elbows, respectively. Set the positions of the elbows considering how different the direction of the arms and legs is. Make sure whether the elbows are attached to the torso. The distance between the left elbow and the left shoulder should be similar to the half of the distance between the left hip and the left knee. The same should apply to the right side as well. 
10. Considering the position, tilt, and angle of the arms of a person taking a specific pose, generate the position of the wrists so that they are well connected to the elbows. It must be generated so that it is clear whether the x-axis, y-axis, and z-axis coordinates are smaller or larger than those of the shoulders, hips, and elbows, respectively. The distance between the left wrist and the left elbow should be similar to the distance between the left elbow and the left shoulder. The same should apply to the right side as well.
11. Considering the human body proportion, generate the position of the nose, eyes, and ears.
12. Format your Keypoints result as shown in the examples. Do not use hash comments. Fill in the coordinates of each keypoint with numbers.

##### Examples (Format)  #####
#     User prompt: A person is doing something.
#     Objects: ['person #1']  
#     Actions: [('person #1', 'doing something')]  
#     Analysis: [('person #1', ['The left hip is ...', 'The right hip is ...', 'The torso is ...', 'The left shoulder is ...', 'The right shoulder is ...', 'The left knee is ...', 'The right knee is ...', 'The left ankle is ...', 'The right ankle is ...', 'The left elbow is ...', 'The right elbow is ...', 'The left wrist is ...', 'The right wrist is ...'])]  
#     Keypoints: [('person #1', {'Nose': [x0, y0, z0], 'Left Eye': [x1, y1, z1], 'Right Eye': [x2, y2, z2], 'Left Ear': [x3, y3, z3], 'Right Ear': [x4, y4, z4], 'Left Shoulder': [x5, y5, z5], 'Right Shoulder': [x6, y6, z6], 'Left Elbow': [x7, y7, z7], 'Right Elbow': [x8, y8, z8], 'Left Wrist': [x9, y9, z9], 'Right Wrist': [x10, y10, z10], 'Left Hip': [x11, y11, z11], 'Right Hip': [x12, y12, z12], 'Left Knee': [x13, y13, z13], 'Right Knee': [x14, y14, z14], 'Left Ankle': [x15, y15, z15], 'Right Ankle': [x16, y16, z16]})]

Your Current Task: Follow the steps closely and accurately identify objects based on the given prompt. Ensure adherence to the above output format.

\end{lstlisting}

\lstset{
    frame=single,                 
    breaklines=true,  
    basicstyle=\footnotesize,
    showstringspaces=False   
}

\subsection{First feedback template}
\label{code:1st_feedback}
\begin{lstlisting}
# Your Role: Excellent Keypoints Adjuster
## Objective: Manipulate keypoints in square images according to the user prompt while maintaining visual accuracy
## Keypoints specification and Manipulations 

1. Keypoints format: [[x0, y0, z0], #Nose
[x1, y1, z1], # Left Eye
[x2, y2, z2], # Right Eye
[x3, y3, z3], # Left Ear
[x4, y4, z4], # Right Ear
[x5, y5, z5], # Left Shoulder
[x6, y6, z6], # Right Shoulder
[x7, y7, z7], # Left Elbow
[x8, y8, z8], # Right Elbow
[x9, y9, z9], # Left Wrist
[x10, y10, z10], # Right Wrist
[x11, y11, z11], # Left Hip
[x12, y12, z12], # Right Hip
[x13, y13, z13], # Left Knee
[x14, y14, z14], # Right Knee
[x15, y15, z15], # Left Ankle
[x16, y16, z16], # Right Ankle
]
Operation: human body action

## Key Guidelines
0. Analyze: Confirm the position of each keypoint individually to describe the movement among the prompt and Analyze the movement. Consider the hips and sit bones as the same.
1. The "special points": Among the keypoints, find the points that are in contact with the ground. We denote that points as "special points". 
2. Alignment: Follow the user's prompt, keeping the specified movement 
3. The ground is xy plane and a person is located in a space where z >0. The z coordinates of the "special points" must be zero and z coordinates of the other points must be greater than or equal to 0.
4. Human body proportion: The following rules must be adhered to without exception. The upper leg length is 0.6 times the back length. The lower leg length is 0.7 times the back length. The upper arm length is 0.3 times the back length. The lower arm length is 0.3 times the back length. The distance from the midpoint of the shoulders to the nose is 0.2 times the back length.

## Process Steps
1. Interpret prompts: Read and understand the user's prompt. Extract the key features for each action accurately. Make the of the hips identical to the position of the sit bones.
2. Find all the "special points". If there are no "special points", set the point closest to the ground as the "special point". The z-coordinate of the "special points" must be exactly 0.
3. Analyze the pose of each object and organize the information about the found features, stating "This keypoint should be in this position. Format your analysis results as shown in the examples.
4. Generate the position of the hips first. The distance between the left hip and the right hip should be small.
5. Determine the position of the shoulders based on whether the torso is standing or lying down while adopting the pose. Determine whether the torso is lying down or not. Make sure to confirm the position of the torso clearly. If the pose is a not a lying down posture, the change in the z-axis between the shoulders and hips should be significant, while the x and y coordinates of the two points should be set to the same value.
6. Considering the fact that a person taking a specific pose is looking to the left, generate the position of the head.
7. Considering the position, tilt, and angle of the legs of a person taking a specific pose, generate the position of the knees so that they are well connected to the hips. It must be generated so that it is clear whether the x-axis, y-axis and z-axis coordinates are smaller or larger than those of the shoulders and hips, respectively. Make sure whether both legs are separated from each other or not. Check the distance between the left knee and the left hip. It should be same as the distance between the right knee and the right hip.
8. Considering the position, tilt, and angle of the legs of a person taking a specific pose, generate the position of the ankles so that they are well connected to the knees. It must be generated so that it is clear whether the x-axis, y-axis, and z-axis coordinates are smaller or larger than those of the shoulders, hips, and knees, respectively. The distance between the left knee and the left ankle should be similar to the distance between the left hip and the left knee. The same should apply to the right side as well.
9. Considering the position, tilt, and angle of the arms of a person taking a specific pose, generate the position of the elbows so that they are well connected to the shoulders. It must be generated so that it is clear whether the x-axis, y-axis, and z-axis coordinates are smaller or larger than those of the shoulders, hips, knees, and elbows, respectively. Set the positions of the elbows considering how different the direction of the arms and legs is. Make sure whether the elbows are attached to the torso. The distance between the left elbow and the left shoulder should be similar to the half of the distance between the left hip and the left knee. The same should apply to the right side as well. 
10. Considering the position, tilt, and angle of the arms of a person taking a specific pose, generate the position of the wrists so that they are well connected to the elbows. It must be generated so that it is clear whether the x-axis, y-axis, and z-axis coordinates are smaller or larger than those of the shoulders, hips, and elbows, respectively. The distance between the left wrist and the left elbow should be similar to the distance between the left elbow and the left shoulder. The same should apply to the right side as well.
11. Considering the human body proportion, generate the position of the nose, eyes, and ears.
12. Format your Keypoints result as shown in the examples. Do not use hash comments. Fill in the coordinates of each keypoint with numbers.

# ## Examples (Format)
#     Objects: ['person #1']  
#     Actions: [('person #1', 'doing something')]  
#     Analysis: [('person #1', ['The left hip is ...', 'The right hip is ...', 'The torso is ...', 'The left shoulder is ...', 'The right shoulder is ...', 'The left knee is ...', 'The right knee is ...', 'The left ankle is ...', 'The right ankle is ...', 'The left elbow is ...', 'The right elbow is ...', 'The left wrist is ...', 'The right wrist is ...'])]  
#     Keypoints: [('person #1', {'Nose': [x0, y0, z0], 'Left Eye': [x1, y1, z1], 'Right Eye': [x2, y2, z2], 'Left Ear': [x3, y3, z3], 'Right Ear': [x4, y4, z4], 'Left Shoulder': [x5, y5, z5], 'Right Shoulder': [x6, y6, z6], 'Left Elbow': [x7, y7, z7], 'Right Elbow': [x8, y8, z8], 'Left Wrist': [x9, y9, z9], 'Right Wrist': [x10, y10, z10], 'Left Hip': [x11, y11, z11], 'Right Hip': [x12, y12, z12], 'Left Knee': [x13, y13, z13], 'Right Knee': [x14, y14, z14], 'Left Ankle': [x15, y15, z15], 'Right Ankle': [x16, y16, z16]})]
#     Answer: ['Yes'/'No']
#     Reason: ['...']

Your Current Task: The following keypoints_list contains the keypoints generated for each object according to the prompt. Determine whether the points in the keypoints_list accurately represent the prompt. If correct, output ["Yes"] as the Answer; if not, output ["No"]. If you output ["Yes"], include the existing action list and analysis list. If you output ["No"], the action_list and analysis_list are the results of analyzing the action and its characteristics for each object based on the user prompt. For each action, correctly modify the analysis and regenerate the keypoints according to the above steps. If you modify the analysis, explain the reason in the Reason section.
Follow the steps closely and accurately identify objects based on the given prompt. Ensure adherence to the above output format.
\end{lstlisting}

\lstset{
    frame=single,                   
    breaklines=true,  
    basicstyle=\footnotesize,
    showstringspaces=False   
}

\subsection{Second feedback template}
\label{code:2nd_feedback}
\begin{lstlisting}
#
Is this depicting '{prompt}'? Answer 'yes' or 'no' and the reason.
#
\end{lstlisting}

\bibliographystyle{elsarticle-num} 
\bibliography{main_revision}

@String(AAAI = {AAAI})

@inproceedings{radford2021learning,
  title={Learning transferable visual models from natural language supervision},
  author={Radford, Alec and Kim, Jong Wook and Hallacy, Chris and Ramesh, Aditya and Goh, Gabriel and Agarwal, Sandhini and Sastry, Girish and Askell, Amanda and Mishkin, Pamela and Clark, Jack and others},
  booktitle={International conference on machine learning},
  pages={8748--8763},
  year={2021},
  organization={PmLR}
}

@inproceedings{zhang2024long,
  title={Long-clip: Unlocking the long-text capability of clip},
  author={Zhang, Beichen and Zhang, Pan and Dong, Xiaoyi and Zang, Yuhang and Wang, Jiaqi},
  booktitle={European Conference on Computer Vision},
  pages={310--325},
  year={2024},
  organization={Springer}
}

@inproceedings{he2016deep,
  title={Deep residual learning for image recognition},
  author={He, Kaiming and Zhang, Xiangyu and Ren, Shaoqing and Sun, Jian},
  booktitle={Proceedings of the IEEE conference on computer vision and pattern recognition},
  pages={770--778},
  year={2016}
}

@inproceedings{verma2020yoga,
  title={Yoga-82: a new dataset for fine-grained classification of human poses},
  author={Verma, Manisha and Kumawat, Sudhakar and Nakashima, Yuta and Raman, Shanmuganathan},
  booktitle={Proceedings of the IEEE/CVF conference on computer vision and pattern recognition workshops},
  pages={1038--1039},
  year={2020}
}

@inproceedings{lin2024evaluating,
  title={Evaluating text-to-visual generation with image-to-text generation},
  author={Lin, Zhiqiu and Pathak, Deepak and Li, Baiqi and Li, Jiayao and Xia, Xide and Neubig, Graham and Zhang, Pengchuan and Ramanan, Deva},
  booktitle={European Conference on Computer Vision},
  pages={366--384},
  year={2024},
  organization={Springer}
}

@article{hessel2021clipscore,
  title={Clipscore: A reference-free evaluation metric for image captioning},
  author={Hessel, Jack and Holtzman, Ari and Forbes, Maxwell and Bras, Ronan Le and Choi, Yejin},
  journal={arXiv preprint arXiv:2104.08718},
  year={2021}
}

@inproceedings{sd,
  title={High-resolution image synthesis with latent diffusion models},
  author={Rombach, Robin and Blattmann, Andreas and Lorenz, Dominik and Esser, Patrick and Ommer, Bj{\"o}rn},
  booktitle={Proceedings of the IEEE/CVF conference on computer vision and pattern recognition},
  pages={10684--10695},
  year={2022}
}

@article{saharia2022photorealistic,
  title={Photorealistic text-to-image diffusion models with deep language understanding},
  author={Saharia, Chitwan and Chan, William and Saxena, Saurabh and Li, Lala and Whang, Jay and Denton, Emily L and Ghasemipour, Kamyar and Gontijo Lopes, Raphael and Karagol Ayan, Burcu and Salimans, Tim and others},
  journal={Advances in neural information processing systems},
  volume={35},
  pages={36479--36494},
  year={2022}
}

@inproceedings{nichol2022glide,
  title={GLIDE: Towards Photorealistic Image Generation and Editing with Text-Guided Diffusion Models},
  author={Nichol, Alexander Quinn and Dhariwal, Prafulla and Ramesh, Aditya and Shyam, Pranav and Mishkin, Pamela and Mcgrew, Bob and Sutskever, Ilya and Chen, Mark},
  booktitle={International Conference on Machine Learning},
  pages={16784--16804},
  year={2022},
  organization={PMLR}
}

@inproceedings{gligen,
  title={Gligen: Open-set grounded text-to-image generation},
  author={Li, Yuheng and Liu, Haotian and Wu, Qingyang and Mu, Fangzhou and Yang, Jianwei and Gao, Jianfeng and Li, Chunyuan and Lee, Yong Jae},
  booktitle={Proceedings of the IEEE/CVF conference on computer vision and pattern recognition},
  pages={22511--22521},
  year={2023}
}

@inproceedings{yang2023reco,
  title={Reco: Region-controlled text-to-image generation},
  author={Yang, Zhengyuan and Wang, Jianfeng and Gan, Zhe and Li, Linjie and Lin, Kevin and Wu, Chenfei and Duan, Nan and Liu, Zicheng and Liu, Ce and Zeng, Michael and others},
  booktitle={Proceedings of the IEEE/CVF Conference on Computer Vision and Pattern Recognition},
  pages={14246--14255},
  year={2023}
}

@article{feng2023layoutgpt,
  title={Layoutgpt: Compositional visual planning and generation with large language models},
  author={Feng, Weixi and Zhu, Wanrong and Fu, Tsu-jui and Jampani, Varun and Akula, Arjun and He, Xuehai and Basu, Sugato and Wang, Xin Eric and Wang, William Yang},
  journal={Advances in Neural Information Processing Systems},
  volume={36},
  pages={18225--18250},
  year={2023}
}

@article{lian2023llm,
  title={Llm-grounded diffusion: Enhancing prompt understanding of text-to-image diffusion models with large language models},
  author={Lian, Long and Li, Boyi and Yala, Adam and Darrell, Trevor},
  journal={arXiv preprint arXiv:2305.13655},
  year={2023}
}

@article{lin2023videodirectorgpt,
  title={Videodirectorgpt: Consistent multi-scene video generation via llm-guided planning},
  author={Lin, Han and Zala, Abhay and Cho, Jaemin and Bansal, Mohit},
  journal={arXiv preprint arXiv:2309.15091},
  year={2023}
}

@inproceedings{wu2024self,
  title={Self-correcting llm-controlled diffusion models},
  author={Wu, Tsung-Han and Lian, Long and Gonzalez, Joseph E and Li, Boyi and Darrell, Trevor},
  booktitle={Proceedings of the IEEE/CVF Conference on Computer Vision and Pattern Recognition},
  pages={6327--6336},
  year={2024}
}

@article{gani2023llm,
  title={Llm blueprint: Enabling text-to-image generation with complex and detailed prompts},
  author={Gani, Hanan and Bhat, Shariq Farooq and Naseer, Muzammal and Khan, Salman and Wonka, Peter},
  journal={arXiv preprint arXiv:2310.10640},
  year={2023}
}

@article{baltruvsaitis2018multimodal,
  title={Multimodal machine learning: A survey and taxonomy},
  author={Baltru{\v{s}}aitis, Tadas and Ahuja, Chaitanya and Morency, Louis-Philippe},
  journal={IEEE transactions on pattern analysis and machine intelligence},
  volume={41},
  number={2},
  pages={423--443},
  year={2018},
  publisher={IEEE}
}

@article{ma2023towards,
  title={Towards local visual modeling for image captioning},
  author={Ma, Yiwei and Ji, Jiayi and Sun, Xiaoshuai and Zhou, Yiyi and Ji, Rongrong},
  journal={Pattern Recognition},
  volume={138},
  pages={109420},
  year={2023},
  publisher={Elsevier}
}

@article{ma2024image,
  title={Image captioning via dynamic path customization},
  author={Ma, Yiwei and Ji, Jiayi and Sun, Xiaoshuai and Zhou, Yiyi and Hong, Xiaopeng and Wu, Yongjian and Ji, Rongrong},
  journal={IEEE Transactions on Neural Networks and Learning Systems},
  year={2024},
  publisher={IEEE}
}

@inproceedings{qu2023layoutllm,
  title={Layoutllm-t2i: Eliciting layout guidance from llm for text-to-image generation},
  author={Qu, Leigang and Wu, Shengqiong and Fei, Hao and Nie, Liqiang and Chua, Tat-Seng},
  booktitle={Proceedings of the 31st ACM International Conference on Multimedia},
  pages={643--654},
  year={2023}
}

@article{fei2024video,
  title={Video-of-thought: Step-by-step video reasoning from perception to cognition},
  author={Fei, Hao and Wu, Shengqiong and Ji, Wei and Zhang, Hanwang and Zhang, Meishan and Lee, Mong-Li and Hsu, Wynne},
  journal={arXiv preprint arXiv:2501.03230},
  year={2024}
}

@article{fei2024enhancing,
  title={Enhancing video-language representations with structural spatio-temporal alignment},
  author={Fei, Hao and Wu, Shengqiong and Zhang, Meishan and Zhang, Min and Chua, Tat-Seng and Yan, Shuicheng},
  journal={IEEE Transactions on Pattern Analysis and Machine Intelligence},
  year={2024},
  publisher={IEEE}
}

@inproceedings{fei2024dysen,
  title={Dysen-vdm: Empowering dynamics-aware text-to-video diffusion with llms},
  author={Fei, Hao and Wu, Shengqiong and Ji, Wei and Zhang, Hanwang and Chua, Tat-Seng},
  booktitle={Proceedings of the IEEE/CVF Conference on Computer Vision and Pattern Recognition},
  pages={7641--7653},
  year={2024}
}

@inproceedings{wu2024next,
  title={NExT-GPT: any-to-any multimodal LLM},
  author={Wu, Shengqiong and Fei, Hao and Qu, Leigang and Ji, Wei and Chua, Tat-Seng},
  booktitle={Proceedings of the 41st International Conference on Machine Learning},
  pages={53366--53397},
  year={2024}
}

@inproceedings{controlnet,
  title={Adding conditional control to text-to-image diffusion models},
  author={Zhang, Lvmin and Rao, Anyi and Agrawala, Maneesh},
  booktitle={Proceedings of the IEEE/CVF international conference on computer vision},
  pages={3836--3847},
  year={2023}
}

@inproceedings{t2i,
  title={T2I-Adapter: learning adapters to dig out more controllable ability for text-to-image diffusion models},
  author={Mou, Chong and Wang, Xintao and Xie, Liangbin and Wu, Yanze and Zhang, Jian and Qi, Zhongang and Shan, Ying},
  booktitle={Proceedings of the Thirty-Eighth AAAI Conference on Artificial Intelligence and Thirty-Sixth Conference on Innovative Applications of Artificial Intelligence and Fourteenth Symposium on Educational Advances in Artificial Intelligence},
  pages={4296--4304},
  year={2024}
}

@inproceedings{humansd,
  title={Humansd: A native skeleton-guided diffusion model for human image generation},
  author={Ju, Xuan and Zeng, Ailing and Zhao, Chenchen and Wang, Jianan and Zhang, Lei and Xu, Qiang},
  booktitle={Proceedings of the IEEE/CVF International Conference on Computer Vision},
  pages={15988--15998},
  year={2023}
}

@article{sdxl,
  title={Sdxl: Improving latent diffusion models for high-resolution image synthesis},
  author={Podell, Dustin and English, Zion and Lacey, Kyle and Blattmann, Andreas and Dockhorn, Tim and M{\"u}ller, Jonas and Penna, Joe and Rombach, Robin},
  journal={arXiv preprint arXiv:2307.01952},
  year={2023}
}

@misc{OpenAI2024o1,
  author    = {OpenAI},
  title     = {Introducing OpenAI o1 Preview},
  year      = {2024},
  url       = {https://openai.com/index/introducing-openai-o1-preview/},
  note      = {Accessed: March 7, 2025}
}

@misc{OpenAI2024o3,
  author    = {OpenAI},
  title     = {Introducing OpenAI o3 and o4-mini},
  year      = {2025},
  url       = {https://openai.com/index/introducing-o3-and-o4-mini/},
  note      = {Accessed: July 11, 2025}
}

@misc{gemini2025_2.5pro,
  author    = {Google DeepMind},
  title     = {Gemini 2.5 Pro},
  year      = {2025},
  url       = {https://deepmind.google/models/gemini/pro/},
  note      = {Accessed: July 11, 2025}
}

@article{cao2019openpose,
  title={Openpose: Realtime multi-person 2d pose estimation using part affinity fields},
  author={Cao, Zhe and Hidalgo, Gines and Simon, Tomas and Wei, Shih-En and Sheikh, Yaser},
  journal={IEEE transactions on pattern analysis and machine intelligence},
  volume={43},
  number={1},
  pages={172--186},
  year={2019},
  publisher={IEEE}
}

@inproceedings{simon2017hand,
  title={Hand keypoint detection in single images using multiview bootstrapping},
  author={Simon, Tomas and Joo, Hanbyul and Matthews, Iain and Sheikh, Yaser},
  booktitle={Proceedings of the IEEE conference on Computer Vision and Pattern Recognition},
  pages={1145--1153},
  year={2017}
}

@inproceedings{cao2017realtime,
  title={Realtime multi-person 2d pose estimation using part affinity fields},
  author={Cao, Zhe and Simon, Tomas and Wei, Shih-En and Sheikh, Yaser},
  booktitle={Proceedings of the IEEE conference on computer vision and pattern recognition},
  pages={7291--7299},
  year={2017}
}

@inproceedings{wei2016convolutional,
  title={Convolutional pose machines},
  author={Wei, Shih-En and Ramakrishna, Varun and Kanade, Takeo and Sheikh, Yaser},
  booktitle={Proceedings of the IEEE conference on Computer Vision and Pattern Recognition},
  pages={4724--4732},
  year={2016}
}

@inproceedings{ruiz2023dreambooth,
  title={Dreambooth: Fine tuning text-to-image diffusion models for subject-driven generation},
  author={Ruiz, Nataniel and Li, Yuanzhen and Jampani, Varun and Pritch, Yael and Rubinstein, Michael and Aberman, Kfir},
  booktitle={Proceedings of the IEEE/CVF conference on computer vision and pattern recognition},
  pages={22500--22510},
  year={2023}
}

@misc{openai_gpt4o,
    title = {Hello GPT-4},
    author = {OpenAI},
    year = {2024},
    url = {https://openai.com/index/hello-gpt-4o/}
}

@article{hao2023optimizing,
  title={Optimizing prompts for text-to-image generation},
  author={Hao, Yaru and Chi, Zewen and Dong, Li and Wei, Furu},
  journal={Advances in Neural Information Processing Systems},
  volume={36},
  pages={66923--66939},
  year={2023}
}

@inproceedings{mo2024dynamic,
  title={Dynamic prompt optimizing for text-to-image generation},
  author={Mo, Wenyi and Zhang, Tianyu and Bai, Yalong and Su, Bing and Wen, Ji-Rong and Yang, Qing},
  booktitle={Proceedings of the IEEE/CVF Conference on Computer Vision and Pattern Recognition},
  pages={26627--26636},
  year={2024}
}

@article{gpt2,
  title={Language models are unsupervised multitask learners},
  author={Radford, Alec and Wu, Jeffrey and Child, Rewon and Luan, David and Amodei, Dario and Sutskever, Ilya and others},
  journal={OpenAI blog},
  volume={1},
  number={8},
  pages={9},
  year={2019}
}

@article{gpt3,
  title={Language models are few-shot learners},
  author={Brown, Tom and Mann, Benjamin and Ryder, Nick and Subbiah, Melanie and Kaplan, Jared D and Dhariwal, Prafulla and Neelakantan, Arvind and Shyam, Pranav and Sastry, Girish and Askell, Amanda and others},
  journal={Advances in neural information processing systems},
  volume={33},
  pages={1877--1901},
  year={2020}
}

@article{t5,
  title={Exploring the limits of transfer learning with a unified text-to-text transformer},
  author={Raffel, Colin and Shazeer, Noam and Roberts, Adam and Lee, Katherine and Narang, Sharan and Matena, Michael and Zhou, Yanqi and Li, Wei and Liu, Peter J},
  journal={Journal of machine learning research},
  volume={21},
  number={140},
  pages={1--67},
  year={2020}
}

@misc{wu2025qwenimagetechnicalreport,
      title={Qwen-Image Technical Report}, 
      author={Chenfei Wu and Jiahao Li and Jingren Zhou and Junyang Lin and Kaiyuan Gao and Kun Yan and Sheng-ming Yin and Shuai Bai and Xiao Xu and Yilei Chen and Yuxiang Chen and Zecheng Tang and Zekai Zhang and Zhengyi Wang and An Yang and Bowen Yu and Chen Cheng and Dayiheng Liu and Deqing Li and Hang Zhang and Hao Meng and Hu Wei and Jingyuan Ni and Kai Chen and Kuan Cao and Liang Peng and Lin Qu and Minggang Wu and Peng Wang and Shuting Yu and Tingkun Wen and Wensen Feng and Xiaoxiao Xu and Yi Wang and Yichang Zhang and Yongqiang Zhu and Yujia Wu and Yuxuan Cai and Zenan Liu},
      year={2025},
      eprint={2508.02324},
      archivePrefix={arXiv},
      primaryClass={cs.CV},
      url={https://arxiv.org/abs/2508.02324}, 
}

@misc{labs2025flux1kontextflowmatching,
      title={FLUX.1 Kontext: Flow Matching for In-Context Image Generation and Editing in Latent Space},
      author={Black Forest Labs and Stephen Batifol and Andreas Blattmann and Frederic Boesel and Saksham Consul and Cyril Diagne and Tim Dockhorn and Jack English and Zion English and Patrick Esser and Sumith Kulal and Kyle Lacey and Yam Levi and Cheng Li and Dominik Lorenz and Jonas Müller and Dustin Podell and Robin Rombach and Harry Saini and Axel Sauer and Luke Smith},
      year={2025},
      eprint={2506.15742},
      archivePrefix={arXiv},
      primaryClass={cs.GR},
      url={https://arxiv.org/abs/2506.15742},
}

@misc{flux2024,
    author={Black Forest Labs},
    title={FLUX},
    year={2024},
    howpublished={\url{https://github.com/black-forest-labs/flux}},
}

@inproceedings{esser2024scaling,
  title={Scaling Rectified Flow Transformers for High-Resolution Image Synthesis},
  author={Esser, Patrick and Kulal, Sumith and Blattmann, Andreas and Entezari, Rahim and M{\"u}ller, Jonas and Saini, Harry and Levi, Yam and Lorenz, Dominik and Sauer, Axel and Boesel, Frederic and others},
  booktitle={International Conference on Machine Learning},
  pages={12606--12633},
  year={2024},
  organization={PMLR}
}

\end{document}